\documentclass[letterpaper]{article} 
\usepackage{aaai24}  
\usepackage{times}  
\usepackage{helvet}  
\usepackage{courier}  
\usepackage[hyphens]{url}  
\usepackage{graphicx} 
\usepackage{natbib}  
\usepackage{caption} 
\usepackage{algorithm}
\usepackage{algorithmic}
\usepackage{newfloat}
\usepackage{listings}
\usepackage{bibentry}
\usepackage{bm}
\usepackage{subcaption}
\usepackage{amssymb}
\usepackage{amsfonts}
\usepackage{amsthm}
\usepackage{amsmath}
\usepackage{booktabs}
\usepackage{multirow}
\usepackage{adjustbox}
\newtheorem{lemma}{Lemma}
\newtheorem{theorem}{Theorem}

\usepackage{newfloat}
\usepackage{listings}
\DeclareCaptionStyle{ruled}{labelfont=normalfont,labelsep=colon,strut=off} 
\floatstyle{ruled}
\newfloat{listing}{tb}{lst}{}
\floatname{listing}{Listing}
\title{ASWT-SGNN: Adaptive Spectral Wavelet Transform-based \\Self-Supervised Graph Neural Network}
\author {
    Ruyue Liu\textsuperscript{\rm 1,\rm 2},
    Rong Yin\textsuperscript{\rm 1,\rm 2}\thanks{Corresponding author.},
    Yong Liu\textsuperscript{\rm 3},
    Weiping Wang \textsuperscript{\rm 1}
}
\affiliations {
    \textsuperscript{\rm 1}Institute of Information Engineering,
Chinese Academy of Sciences\\
    \textsuperscript{\rm 2}University of Chinese Academy of Sciences\\
    \textsuperscript{\rm 3}Renmin University of China\\
    \{liuruyue, yinrong, wangweiping\}@iie.ac.cn, liuyonggsai@ruc.edu.cn
}

\usepackage{bibentry}

\begin{document}

\maketitle

\begin{abstract}
Graph Comparative Learning (GCL) is a self-supervised method that combines the advantages of Graph Convolutional Networks (GCNs) and comparative learning, making it promising for learning node representations. However, the GCN encoders used in these methods rely on the Fourier transform to learn fixed graph representations, which is inherently limited by the uncertainty principle involving spatial and spectral localization trade-offs. To overcome the inflexibility of existing methods and the computationally expensive eigen-decomposition and dense matrix multiplication, this paper proposes an Adaptive Spectral Wavelet Transform-based Self-Supervised Graph Neural Network (ASWT-SGNN). The proposed method employs spectral adaptive polynomials to approximate the filter function and optimize the wavelet using contrast loss. This design enables the creation of local filters in both spectral and spatial domains, allowing flexible aggregation of neighborhood information at various scales and facilitating controlled transformation between local and global information. Compared to existing methods, the proposed approach reduces computational complexity and addresses the limitation of graph convolutional neural networks, which are constrained by graph size and lack flexible control over the neighborhood aspect. Extensive experiments on eight benchmark datasets demonstrate that ASWT-SGNN accurately approximates the filter function in high-density spectral regions, avoiding costly eigen-decomposition. Furthermore, ASWT-SGNN achieves comparable performance to state-of-the-art models in node classification tasks.
\end{abstract}

\section{Introduction}
Graphs are essential in various real-world domains such as social networks, brain networks, transportation networks, and citation networks \cite{wang2016causality,yu2020forecasting}. Recently, the emergence of graph neural networks (GNNs) \cite{he2022block,wang2022uncovering} has attracted much attention due to their success in applications involving graph-structured data such as node classification \cite{welling2016semi} and edge prediction \cite{hasanzadeh2019semi}. Graph Comparative Learning (GCL) \cite{velickovic2019deep, sun2019infograph} combines the capabilities of GNN and comparative learning techniques, making it a promising paradigm in the field of graph analysis \cite{he2020momentum}. Typically, GCL methods generate multiple views by randomly augmenting input data and optimize the GNN encoder by learning consistency across views. GCL reduces the dependence of graph representation learning on human annotations and achieves state-of-the-art performance in tasks such as node classification \cite{hassani2020contrastive, zhang2021canonical}.

Most graph contrastive learning methods utilize graph convolutional neural networks (GCNs) as encoders \cite{xie2022self, wang2022uncovering}. Similar to convolutional neural networks (CNNs) in computer vision, spectral GCNs use Fourier bases in the design of graph-based operators \cite{bruna2014spectral}. However, these operators are localized in the frequency rather than the spatial domain. Additionally, they require costly multiplications between eigen-decomposition and dense matrices, leading to high computational expenses. In order to address this issue and achieve spatial localization, methods such as ChebyNet \cite{defferrard2016convolutional} and GCN \cite{welling2016semi} employ polynomial approximation. While GCN is widely adopted for graph problems due to its impressive performance and computational efficiency \cite{shi2020multi}, it encounters limitations and challenges when applied to large graphs, especially in mini-batch settings \cite{zeng2019graphsaint}. To overcome the scaling challenges of GCN on large graphs, researchers have proposed layer sampling methods \cite{ying2018hierarchical, kaler2022accelerating} and subgraph sampling methods \cite{zeng2019graphsaint, zeng2019accurate}. However, the filter size is determined by the size of the entire graph or the sampled subgraph, which restricts flexibility for inputs of different sizes. Although some flexible spatial methods have been proposed, their aggregators lack learnability and convolutional properties. Consequently, the problem of designing a flexible filter that combines the learnability of spatial methods and the convolutional properties of spectral methods still needs to be solved.

To address the issues of inflexible and unlearnable filters, as well as the limited applicability caused by high computational complexity in existing methods, this paper proposes a novel graph comparative learning paradigm based on the adaptive spectral wavelet transform. More specifically, a fast spectral adaptive approximation method is utilized to estimate the wavelet filter, and contrast loss is employed to optimize the wavelet scale directly. Additionally, the introduction of residual links mitigates over-smoothing during information aggregation. By avoiding the expensive eigen-decomposition of the graph Laplacian operator and enabling localization in both the spectral and spatial domains, this approach effectively overcomes the limitations of graph convolutional neural networks that are constrained by graph size and lack flexibility in controlling neighborhood aspects. In comparison to state-of-the-art methods, this work introduces the following innovations:

\begin{itemize}
    \item We propose a novel self-supervised graph representation learning method based on sparse graph wavelets that creates localized filters in both the spectral and spatial domains. It reduces the computational complexity and addresses the limitation that graph convolutional neural networks cannot flexibly control the neighborhood aspect.
    \item Theoretically, we demonstrate that nodes with similar network neighborhoods and features exhibit similar ASWT-SGNN embeddings, providing a performance guarantee for the proposed method. 
    \item Extensive experiments on eight benchmark datasets show that the proposed method reduces approximation errors in high-density spectral components without requiring expensive eigen-decomposition and achieves competitive performance with state-of-the-art models in node classification tasks.
\end{itemize}

\section{Related Works}
\label{related work}
\subsection{Graph Convolutional Neural Network}
Following the success of CNNs in computer vision and natural language processing, researchers have sought to extend CNNs to the graph domain. The key challenge lies in defining the convolution operator for graphs. Graph convolutional neural networks can be broadly categorized into two main approaches: spectral and spatial. Spectral methods employ the graph Fourier transform to transfer signals from the spatial domain to the spectral domain, where convolution operations are performed. Spectral GNN \cite{bruna2014spectral} is the first attempt to implement CNNs on graphs. ChebyNet \cite{defferrard2016convolutional} introduce a parameterization method using Chebyshev polynomials for spectral filters, which enables fast localization of spectral filters. GCN \cite{welling2016semi} proposes a simplified version of ChebyNet, which achieves success in graph semi-supervised learning tasks. However, these spectral methods face challenges with generalization, as they are limited by fixed graph sizes, and larger filter sizes result in increased computational and memory costs. Spatial methods draw inspiration from weighted summation in CNNs to extract spatial features from topological graphs. MoNet (Monti et al. 2017) provides a general framework for designing spatial methods by utilizing the weighted average of functions defined within a node’s neighborhood as a spatial convolution operator. GAT \cite{velickovic2018graph} proposes a self-attentive mechanism to learn the weighting function. However, these methods employ unlearnable aggregators, and the localization of the convolution operation remains uncertain.

\subsection{Graph Contrastive Learning}
In graph analysis, contrastive learning was initially introduced by DGI \cite{velickovic2019deep} and InfoGraph \cite{sun2019infograph}, drawing inspiration from maximizing local-global mutual information. Building upon this, MVGRL \cite{hassani2020contrastive} incorporate node diffusion into the graph comparison framework. GCA \cite{zhu2021graph} learns node representations by considering other nodes as negative samples, while BGRL \cite{thakoor2021bootstrapped} proposes a no-negative-sample model. CCA-SSG \cite{zhang2021canonical} optimizes feature-level objectives in addition to instance-level differences. GRADE \cite{wang2022uncovering} investigates fairness differences in comparative learning and proposes a novel approach to graph enhancement. Several surveys \cite{xie2022self, ding2022data,kumar2022contrastive} summarize recent advancements in graph contrastive learning. Despite these methods' notable achievements, most rely on GCN and its variants as the base models, inheriting the limitations of GCN. These limitations restrict the performance of these graph contrastive learning methods in tasks that require preserving fine-grained node features.

\subsection{Graph Wavelets}
The wavelet transform exhibits favorable structural properties by utilizing finite length and attenuation basis functions. This approach effectively localizes signals within both the spatial and spectral domains. Additionally, it is noteworthy that the basis and its inverse in the wavelet transform often showcase sparsity, contributing to its utility and efficiency. To construct the wavelet transform on graphs, Hammond et al. \cite{hammond2011wavelets} propose a method that approximates the wavelet using Chebyshev polynomials. This approach effectively avoids the need for eigen-decomposition of the Laplace matrix. Building upon this, GWNN \cite{xu2019graph} redefines graph convolution based on graph wavelets, resulting in high efficiency and sparsity. M-GWCN \cite{behmanesh2022geometric} applies the multi-scale graph wavelet transform to learn representations of multimodal data. Collectively, these works showcase the value of graph wavelets in signal processing on graphs. However, these methods primarily employ wavelet transforms in supervised or semi-supervised tasks and heavily rely on labeled data.

\begin{figure*}[t]
\centering
    \begin{subfigure}{0.42\textwidth}
        \centering
        \includegraphics[width=1\textwidth,height=4.3cm]{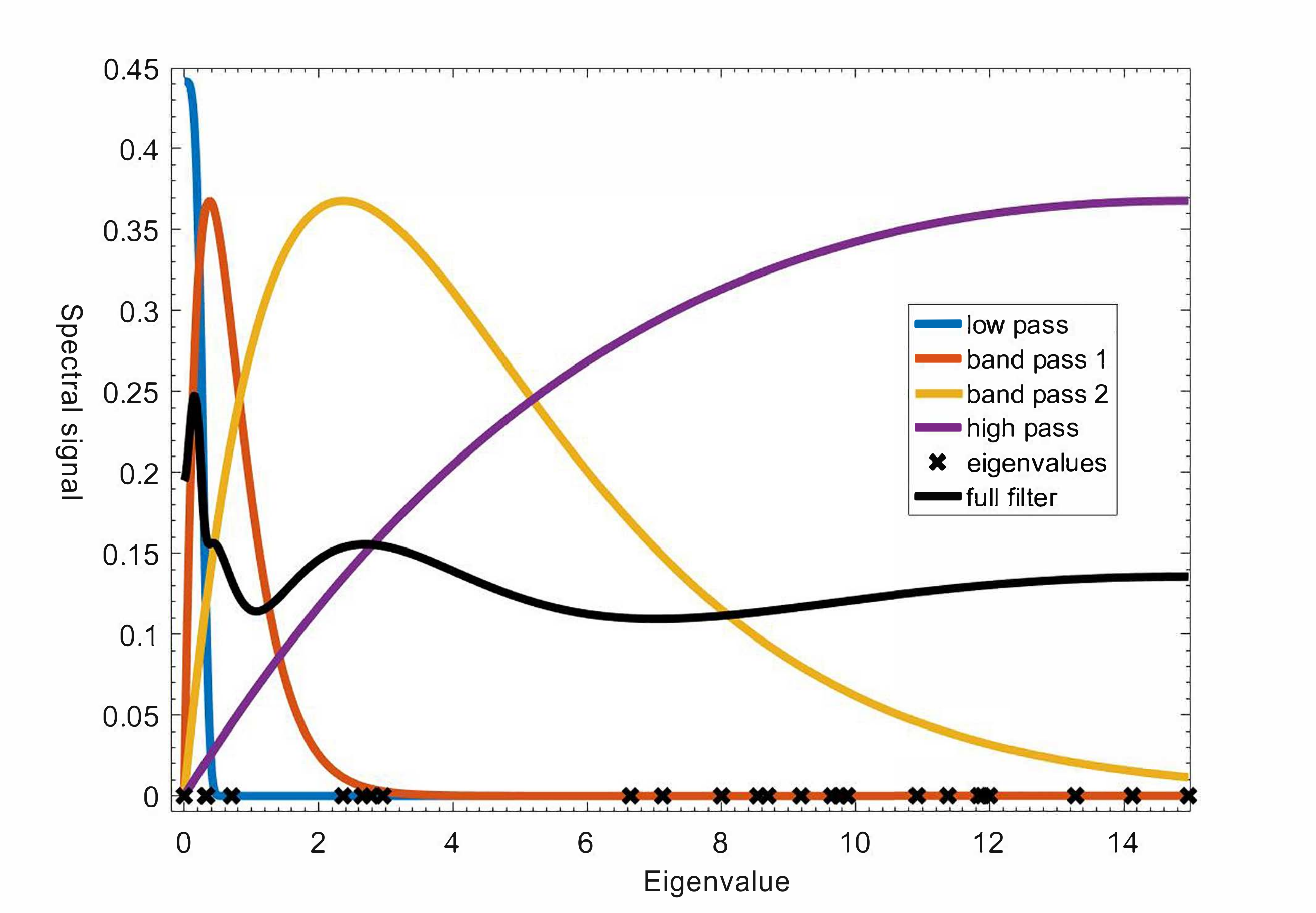}
        \caption{Signal spectrum}
        \label{fig:subfig1}
    \end{subfigure}
    \begin{subfigure}{0.55\textwidth}
        \centering
        \includegraphics[width=1\textwidth,height=4.5cm]{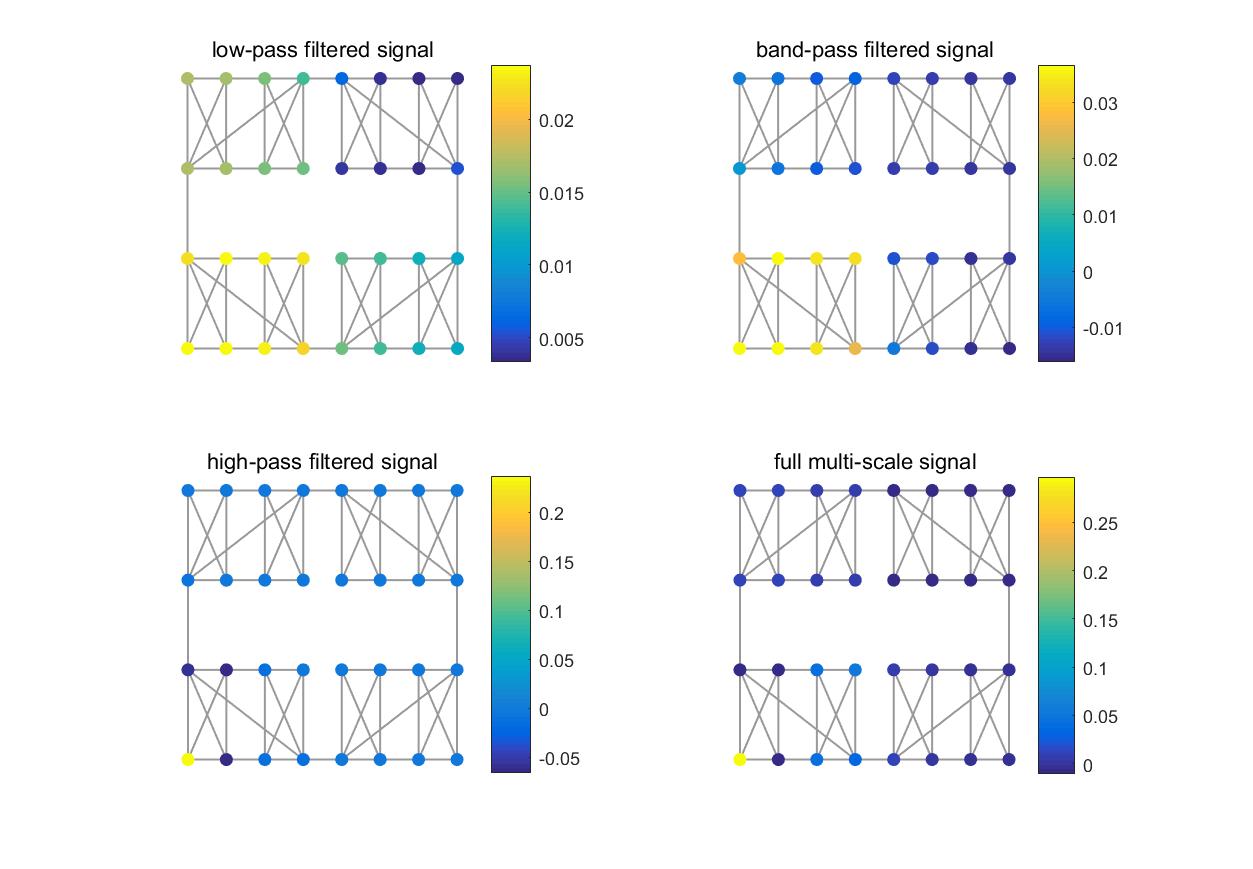}
        \caption{Filter signal}
        \label{fig:subfig2}
    \end{subfigure}
\caption{Visualize how wavelet filters can capture multiscale properties in graph signals and structures. We use graphs with two levels of clusters (4-node clusters and 8-node clusters) for demonstration. These clusters are reflected in the gaps (about 0 and 6) in the spectrum in Figure \ref{fig:subfig1}, reflecting different eigenvalue clustering. The signal is obtained by filtering a random signal with the filter in Figure \ref{fig:subfig1}, purposefully highlighting the three eigenvalue clusters. Figure \ref{fig:subfig2} shows how the complete signal is decomposed into multiple filter components.}
\label{fig1}
\end{figure*}

\section{Preliminary}
\subsection{Graph Fourier Transform}
Let $\mathcal{G} =(\mathcal{V},\mathcal{E})$ denotes a graph, where $\mathcal{V}= \{v_{1},\cdots ,v_{N}\}$ represents the set of nodes and $\mathcal{E} \subseteq \mathcal{V} \times \mathcal{V}$ represents the set of edges. $\mathcal{G}$ is associated with a feature matrix $\bm{X} \in \mathbb{R}^{N \times p}$, $\bm{X}=[x_1, \dots, x_n]$ and an adjacency matrix $\bm{A} \in \mathbb{R}^{N\times N}$, where $\bm{A}_{ij}=1$\, if $(v_{i},v_{j})\in \mathcal{E}$ and $\bm{A}_{ij}=0$\, otherwise. We define the Laplacian matrix of the graph as $\bm{L}=\bm{D}-\bm{A}$, where $\bm{D} = Diag(d_{1},\dots ,d_{N})$, $d_i={\textstyle \sum_{j}^{}\bm{A}_{ij}}$. The symmetric normalized Laplacian is defined as $ \bm{L}_{sym}=\bm{D}^{-\frac{1}{2}} \bm{L} \bm{D}^{-\frac{1}{2}}=\bm{U} \bm{\Lambda} \bm{U}^{\top}$. 
$\bm{U}=\left[{\bm{u}}_{1}, \ldots, {\bm{u}}_{N}\right]$, where $\bm{\bm{u}}_{i} \in \mathbb{R}^{N}$ denotes the $i$-th eigenvector of $\bm{L}_{sym}$ and $\bm{\Lambda} = Diag (\lambda_{1},\dots ,\lambda_{N})$ is the corresponding eigenvalue matrix.

For a discrete graph signal $\bm{X}$, its Fourier transform has the following form \cite{shuman2013emerging}:
\begin{equation}
\hat{f}\left(\lambda_{\ell}\right)=\sum_{i=1}^{N} \bm{x}_i \bm{U}_{\ell}^{*}(i)=\sum_{i=1}^{N} \bm{x}_i \bm{U}_{\ell}^{T}(i),
\label{eq1}
\end{equation}
where $\bm{U}_{\ell}^{*}$ denotes the conjugate transposition of the eigenvalues $\lambda_{\ell}$ corresponding to the eigenvectors. Since the complex number is not involved in the scope of this work, it can be regarded as a common transposition, that is, $\bm{U}^{*}=\bm{U}^{T}$.

Graph Fourier transform provides a means to define the graph convolution operator using the convolution theorem. The filter signal, denoted as $f_o$, can be mathematically expressed as:
\begin{equation}
f_o =\bm{U} g(\bm{\Lambda}) \bm{U}^{T} \bm{X},
\label{eq2}
\end{equation}
where $g(\bm{\Lambda})$ denotes the filter function on the eigenvalues, and $\bm{U} g(\bm{\Lambda}) \bm{U}^{T}$ is the graph filtering matrix.

\subsection{Uncertainty Principle}
The implementation of graph convolution using the Fourier transform lacks spatial localization despite its ability to achieve localization in the spectral domain. Additionally, its localization is significantly influenced by computational efficiency. We employ the spatial and spectral concentration metric proposed by Tsitsvero et al. \cite{tsitsvero2016signals} to establish a more comprehensive and precise notion of localization.

\begin{equation}
\frac{\left\|\bm{B}\bm{x}\right\|_2^2}{\left\|\bm{x}\right\|_2^2}=a^2, \frac{\left\|\bm{C}\bm{x}\right\|_2^2}{\left\|\bm{x}\right\|_2^2}=b^2,
\label{eq3}
\end{equation}
where the square of the Euclidean paradigm $\left\|\bm{x}\right\|_2^2$ denotes the total energy of the signal $\bm{x}$. $\bm{B}$ is the diagonal matrix representing the node restriction operator. Given a node subset $\mathcal{S}\subseteq\mathcal{V}$, $\bm{B}={Diag}\left\{\mathbb{I}(i\in \mathcal{S})\right\}$, where $\mathbb{I}(\cdot)$ is an indicator function. $\bm{C}=\bm{U}\Sigma_{\mathcal{F}}\bm{U}^{-1},$ is the band-limiting operator. Given a matrix $\bm{U}$ and a subset of frequency indices $\mathcal{F} \subseteq \mathcal{V}^*$, where $\mathcal{V}^* = \{1,\cdots, N\}$ denotes the set of all frequency indices. $\Sigma_{\mathcal{F}}$ is a diagonal matrix defined as $\Sigma_{\mathcal{F}} ={Diag}\left\{\mathbb{I}(i\in \mathcal{F})\right\}$.

More specifically, considering a pair of vertex sets $\mathcal{S}$ and frequency sets $\mathcal{F}$, where $a^2$ and $b^2$ represent the percentage of energy contained within the sets $\mathcal{S}$ and $\mathcal{F}$ respectively, our objective is to determine the balance between $a$ and $b$ and identify signals that can achieve all feasible pairs. The resulting uncertainty principle is formulated and presented in the following theorem \cite{tsitsvero2016signals}.

\begin{equation}
\cos^{-1}a+\cos^{-1}b\ge\cos^{-1}\lambda_{\max}\left(\bm{BC}\right),
\label{eq4}
\end{equation}
where $\lambda_{max}(\bm{BC})$ is the maximum eigenvalue of $\bm{BC}$.

Considering the uncertainty principle, we retain the entire frequency band if we follow the graph localization method in Eq. (\ref{eq2}). However, real graph signals show a non-uniform distribution in the frequency domain, which suggests that $\bm{C}$ can be chosen more efficiently. Moreover, a deeper network leads to a wider propagation of the signal in the graph domain, which limits the width of the frequency bands and leads to the dominance of low-frequency signals (smoothing signals), which produce over-smoothing.

\subsection{Graph Wavelet Transform}
As the graph Fourier transform, the graph wavelet transform also necessitates a set of suitable bases to map the graph signal to the spectral domain. In this case, we denote the wavelet operator as $\bm{\Psi}_{s}=\bm{U} g_s(\bm{\Lambda}) \bm{U}^{\top}$, where $s$ is the scaling parameter. The wavelet transform breaks down a function $g$ into a linear combination of basis functions localized in spatial and spectral. This paper employs the Heat Kernel wavelet as the low-pass filter (denoted as $g^{l}$). In contrast, the Mexican-Hat wavelet is the band-pass filter (denoted as $g^{b}$). These filters are defined as follows:
\begin{equation}
g^{l}_{s}(\lambda) = e^{-s\lambda},
\label{eq5}
\end{equation}
\begin{equation}
g_s^b(\lambda)=\frac{2}{\sqrt{3}\pi^{\frac{1}{4}}}\left(1-\left({\lambda}{s}\right)^2\right)e^{-\frac{\left({\lambda}{s}\right)^2}{2}}.
\label{eq6}
\end{equation}

In this work, we integrate filter functions to achieve the combined effect of a low-pass filter and a wide band-pass filter. Figure 1 exemplifies how the combination of low-pass and band-pass filters can generate a more intricate wavelet filter capable of capturing signal components at various scales. This design enables us to obtain a richer representation of the signal, encompassing its frequency information more nuancedly.

\begin{equation}
g_\theta(\lambda)=g_{s_{0}}^{l}(\lambda)+\sum_{l=1}^Lg_{s_{l}}^{b}(\lambda),
\label{eq7}
\end{equation}
where $\theta = \{s_{0}, s_{1}, ..., s_{l}\}$ is the set of scale parameters.

Using the graph wavelet transform instead of the graph Fourier transform in Eq. (\ref{eq2}), we get the graph convolution as follows:
\begin{equation}
f_{o}= \bm{\Psi} \bm{G} \bm{\Psi}^{\top}\bm{X},
\label{eq8}
\end{equation}
where $\bm{G}$ is a diagonal matrix, which acts as a filter, the scale set $\theta$ of wavelet coefficients is omitted for simplification.

\section{Methodology}
\subsection{Wavelet Coefficients Approximation}
The model formulation discussed in the preceding sections necessitates an eigen-decomposition of the Laplace operator of the input graph $\mathcal{G}$. However, this process presents a computational complexity of $\mathcal{O}(N^3)$, rendering it infeasible for larger graphs. In order to overcome this constraint, we employ the polynomial approximation method previously introduced by Hammond et al. \cite{hammond2011wavelets}. This method involves expressing the wavelet filter as $g_{\theta}(\lambda) \approx  p_{\theta}(\lambda) = \gamma_0 + \gamma_1\lambda + \cdots + \gamma_m\lambda^m$. This allows rewriting the wavelet operator as $\bm{\Psi_{\theta}} = \bm{U} p_{\theta}(\bm{\Lambda})\bm{U}^{\top} = p_{\theta}(\bm{L}_{sym})$. 

While existing methods rely on Chebyshev polynomial approximations, we aim to optimize scales. Therefore, we utilize a least squares approximation that can parameterize the set of wavelet scales $\theta$, which can be expressed as follows: 
\begin{equation}
\bm{\gamma}_{\theta}=(\bm{V}_{\bm{\Lambda}}^{\top}\bm{V}_{\bm{\Lambda}})^{-1}\bm{V}_{\bm{\Lambda}}^{\top}g_{\theta}(\bm{\Lambda}),
\label{eq9}
\end{equation}
Where $\bm{V}_{\bm{\Lambda}} \in \mathbb{R}^{N \times (m+1)}$ is the Vandermonde matrix of $\bm{\Lambda}$ from order $0$ to order $m$, $N$ is the number of eigenvalues.

Accurate calculation of the eigenvalues requires an expensive eigen-decomposition of the graph Laplace operator, which is not feasible in our case. As an alternative, we transform the set of eigenvalues into a sequence of linearly spaced points $\bm{\xi} = \{ \xi_i\}_{i=1}^K $ within the interval $[0,2]$ on the spectral domain. However, in graphs that exhibit multiscale features, the eigenvalues do not follow a uniform distribution on the spectral domain. Instead, they display spectral gaps corresponding to different scales within the data. To address this non-uniform distribution, we incorporate the estimated spectral density $\bm{\omega}$ as the weight for each of the $K$ sample points $\bm{\xi}$ on the spectral domain \cite{fan2020spectrum}. Consequently, we can compute the weighted least squares coefficients,
\begin{equation}
\bm{\gamma}_{\theta}=(\bm{V}_{\bm{\xi}}^{\top}Diag(\bm{\omega})\bm{V}_{\bm{\xi}})^{-1}\bm{V}_{\bm{\xi}}^{T}Diag(\bm{\omega})g_{\theta}(\bm{\xi}),
\label{eq10}
\end{equation}
where the spectral density $\bm{\omega} = \{\omega_j\}_{j=1}^K$, and ${\omega}_i = \frac{1}{N}\sum_{j=1}^{N}\{\mathbb{I}(\lambda_{j}=\xi_i)\}$.

The goal of spectral density estimation is to approximate the density function without expensive graph Laplacian eigen-decomposition. To achieve this, we determine the number of eigenvalues less than or equal to each $\xi_i$ in the set $\bm{\xi}$. It can be achieved by computing an approximation to the trace of the eigen-projection matrix $\bm{P}$ \cite{di2016efficient}. In practice, directly obtaining the projector $\bm{P}$ is often not feasible. However, it can be approximated efficiently using polynomials or rational functions of the Laplacian operator $\bm{L}$. In this approximation, we interpret $\bm{P}$ as a step function of $\bm{L}$, which can be expressed as follows:

\begin{equation}
\bm{P}_{\xi_i} = h(\bm{L}),~\text{where}~h(\lambda)=\left\{
\begin{array}{ll}
1, & \text{if}~\lambda\le \xi_i \\
0, & \text{otherwise.}
\end{array}\right.
\label{eq11}
\end{equation}

Although it is impossible to compute $h(\lambda)$ exactly cheaply, it can be approximated using a finite sum of Jackson-Chebyshev polynomials, denoted as $\phi(\lambda)$. Please refer to Appendix A\ref{A} for detailed information on the approximation method. In this form, it becomes possible to estimate the trace of $\bm{P}$ by an estimator developed by Hutchinson \cite{hutchinson1989stochastic} and further improved more recently \cite{tang2012probing}. Hutchinson’s stochastic estimator relies solely on matrix-vector products to approximate the matrix trajectory. The key idea is to utilize Rademacher random variables, where each entry of a randomly generated vector $\bm{R} \in \mathbb{R}^N$ takes on the values $-1$ and $1$ with equal probability of $\frac{1}{2}$. Thus, an estimate of the trace $tr(\bm{P})$ can be obtained by generating $n_r$ samples of random vectors $\bm{R}_k$, $k = 1,\cdots, n_r$ and computing the average over these samples. The estimator can be expressed as follows:
\begin{equation}
{tr}(\bm{P})\approx \frac{1}{n_{r}}\sum_{k=1}^{n_{r}}\bm{R}_{k}^{\top}\bm{P}\bm{R}_{k} \approx \frac{1}{n_{r}}\sum_{k=1}^{n_{r}}\bm{R}_{k}^{\top}\phi({\bm{L}_{sym}})\bm{R}_{k}.
\label{eq12}
\end{equation}

We obtain the approximation $\bm{\Omega}$ to the cumulative spectral density function.
\begin{equation}
\bm{\Omega}=\left\{\left(\xi_{i},\frac{1}{N}\left[\frac{1}{n_{r}}\sum_{k=1}^{n_{r}}\bm{R}_{k}^{\top}\phi({\bm{L}_{sym}})\bm{R}_{k}\right]\right)\right\}_{i=1}^{K}.
\label{eq13}
\end{equation}

Finally, the cumulative spectral density $\bm{\Omega}$ is differentiated to obtain an approximation of the spectral density $\bm{\omega}$, ie $\bm{\omega}=\frac{d}{d\bm{\xi}} \bm{\Omega}$. Using the estimated spectral density $\bm{\omega}$ as a weight for each sample point $\xi_i$ on the spectral domain, the weighted least squares coefficient $\bm{\gamma}_\theta$ can be calculated by substituting it into Eq. (\ref{eq10}).  

These coefficients are used to approximate the wavelet filter matrix $\bm{\Psi}_\theta$, which can be expressed as follows:
\begin{equation}
\bm{\Psi}_\theta = \gamma_0\bm{I}+ \gamma_1\bm{L}_{sym}+\cdots+\gamma_m\bm{L}_{sym}^m.
\label{eq14}
\end{equation}

\subsection{Encoder}
In this paper, we construct a graph wavelet multilayer convolutional network (ASWT-SGNN). Based on the above, we define the $l$-th layer of ASWT-SGNN as
\begin{equation}
\bm{H}^{l+1}=\sigma(\bm{H}^{l^{\prime}}\bm{W}^{l}),
\label{eq15}
\end{equation}
where $\sigma$ is the activation function, $\bm{H}^{l^{\prime}}\in\mathbb{R}^{N \times p}$ is the results of graph convolution of  the $l$-th layer, $\bm{W}^{l} \in \mathbb{R}^{p \times q}$ is the weight of the $l$-th layer, $p$ is the number of features in current layer, and $q$ is the number of features in next layer. $\bm{H}^{l^{\prime}}$ in Eq. (\ref{eq15}) is described as follows:
\begin{equation}
\bm{H}^{{l}^{\prime}}=\alpha\bm{F}^l\bm{H}^{l}+(1-\alpha)\bm{H}^{{l}},
\label{eq16}
\end{equation}
where $1-\alpha$ represents the proportion of the original features $\bm{H}^{l}$ in $l$-th layer, and $0 \le \alpha \le 1$. By incorporating this residual connection, we guarantee that regardless of the number of layers we add, the resulting representation will always retain a portion of the initial features. To enhance the encoder's capability to aggregate both local and global information effectively, we define the diffusion operator $\bm{F}^l$ of $l$-th layer in Eq. (\ref{eq16}) as
\begin{equation}
{\bm{F}^l}=\beta(\bm{\Psi}_\theta\bm{G}^l\bm{\Psi}_{\bm{\theta}}^{\top})+(1-\beta)({\bm{D}}^{-\frac{1}{2}}\tilde{\bm{A}}{\bm{D}}^{-\frac{1}{2}}),
\label{eq17}
\end{equation}
where $\beta$ stands for the ratio of the graph wavelet term and $0 \le \beta \le 1$. ${\theta}$ is the learnable multi-scale parameters. $\bm{G}^{l}$ is the learnable diagonal filter matrix of $l$-th layer. $\tilde{\bm{A}}=\bm{A}+\bm{I}$ represents the adjacency matrix of the self-loop graph of $\mathcal{G}$, where $\bm{I}$ is the identity matrix.

\subsection{Optimization Objective}
Typical GCL methods involve generating augmented views and subsequently optimizing the congruence between their encoded representations. In this paper, we generate two augmented graphs, $\bm{z}$ and $\bm{o}$, by using feature augmentation. We randomly sample the mask vector $\bm{m_f}\in \{0,1\}^B$ to hide part of the dimensions in the node feature. Each element in mask $\bm{m_f}$ is sampled from Bernoulli distribution $Ber(1- f_d)$, where the hyperparameter $f_d$ is the feature descent rate. Therefore, the augmented node feature $\hat{\bm{X}}$ calculated by the following formula:
\begin{equation}
\hat{\bm{X}}=\left[\bm{x}_{1} \circ \bm{m}_f, \bm{x}_{2} \circ \bm{m}_f, \cdots, \bm{x}_{N} \circ \bm{m}_f\right].
\label{eq18}
\end{equation}

We use the comparison objective for the node representation of the two graph augmentation obtained. For node $v_a$, the node representations from different graph augmentation $\bm{z}_a$ and $\bm{o}_a$ form a positive pair. In contrast, the node representations of other nodes in the two graph augmentation are considered negative pairs. Therefore, we define the paired objective of each positive pair $(\bm{z}_a, \bm{o}_a)$ as
\begin{equation}
\mathcal{L}_a(\bm{z},\bm{o}) = \log \frac{e^{\theta\left(\bm{z}_{a}, \bm{o}_{a}\right) }}{e^{\theta\left(\bm{z}_{a}, \bm{o}_{a}\right) }+\sum_{b \neq a}^{} (e^{\theta\left(\bm{z}_{a}, \bm{o}_{b}\right) }+e^{\theta\left(\bm{z}_{a}, \bm{z}_{b}\right)})},
\label{eq19}
\end{equation}
where the critic $\theta(\bm{z},\bm{o})$ is defined as $sim(\bm{H_z},\bm{H_o})$, and $sim(\cdot,\cdot)$ refers to cosine similarity function. $\bm{H_z}$ is the graph embedding generated by graph augmentation through the proposed method ASWT-SGNN. The overall objective to be maximized is the average of all positive pairs,
\begin{equation}
\mathcal{L}=- \frac{1}{2 N} \sum_{a=1}^{N}\left[\mathcal{L}_a\left(\bm{z}, \bm{o}\right)+\mathcal{L}_a\left(\bm{o}, \bm{z}\right)\right].
\label{eq20}
\end{equation}

\subsection{Complexity Analysis}
The first step of ASWT-SGNN  is wavelet coefficients approximation. Eq. (\ref{eq8}) shows that the immediate solution requires eigen-decomposition of the Laplace matrix to obtain the eigenvalues and eigenvectors of the matrix. However, the complexity of the direct solution is very high. For example, the time complexity of the quick response (QR) algorithm is $\mathcal{O}(N^3)$, and the space complexity is $\mathcal{O}(N^2)$. Therefore, we use the least squares approximation to approximate the solution in this step. If the $m$-order Chebyshev polynomial approximation is used, the sum of the complexity of each item in the polynomial is calculated as $\mathcal{O}(m \times |E|)$. 

The second step of ASWT-SGNN is to use the wavelet transform for graph convolution. Spectral CNN \cite{bruna2014spectral} has high parameter complexity $\mathcal{O}(N\times p\times q)$. ChebyNet \cite{defferrard2016convolutional} approximates the convolution kernel by the polynomial function of the diagonal matrix of the Laplace eigenvalue, reducing the parameter complexity to $\mathcal{O}(m \times p\times q)$, where $m$ is the order of the polynomial function. GCN \cite{welling2016semi} simplifies ChebyNet by setting m=1. In this paper, the feature transformation is performed first, and the parameter complexity is $\mathcal{O}(p\times q)$. Then the graph convolution is performed, and the parameter complexity is $\mathcal{O}(N)$. 

\subsection{Theoretical Analysis}
\begin{lemma}
Consider nodes $v_a$ and $v_b$ within a graph, characterized by their similarity in features or labels. Consequently, their $k$-hop neighbors exhibit a one-to-one mapping, specifically $\mathcal{N}_k(b) = \pi(\mathcal{N}_k(a))$. In this context, it holds true that $|\bm{\Psi}_{am} - \bm{\Psi}_{b\pi(m)}| \le 2\epsilon$, where $\bm{\Psi}_{am}$ signifies the wavelet coefficient between nodes $v_a$ and $v_m$.
\label{lem1}
\end{lemma}

\begin{proof} 
The proof is shown in Appendix A\ref{A}. 
\end{proof}

This lemma shows that nodes with similar network neighborhoods have similar wavelet coefficients.

\begin{theorem}
If the graph following the above assumption and Lemma 
\ref{lem1}, the expectation of embedding is given by:
\begin{equation}
\begin{split}
 \mathbb{E}\left[\bm{F}_{i}\right]=\bm{W} \mathbb{E}_{y \sim P\left(y_{i}\right), \bm{x} \sim P_{y}(\bm{x})}[\bm{\Gamma}_i\bm{x}],
\end{split}
\label{eq21}
\end{equation}
where $\bm{\Gamma} = \bm{\Psi} \bm{G} \bm{\Psi}^{\top}$, $\bm{\Gamma}_i$ is the $i$-th row of  $\bm{\Gamma}$. In this context, we simplify the model by excluding the residual connection component. With probability at least $1-\delta $ over the distribution for the graph, we have: 
\begin{equation}
\begin{split}
 \left\|\bm{H}_{i}-\mathbb{E}\left[\bm{H}_{i}\right]\right\|_{2} \leq \sqrt{\frac{\sigma_{\max }^{2}(\bm{W}) p \log (2 p / \delta)}{2 \bm{N}\|\bm{\Gamma}_i\bm{x}\|_{\psi_{2}}}},
\end{split}
\label{eq22}
\end{equation}
where the sub-gaussian norms $\|\bm{\Gamma}_i \bm{x}\|_{\psi_{2}} = \min\left\|\bm{\Gamma}_i\bm{x}_{i, d}\right\|_{\psi_{2}}$, $p$ is the dimension of features, $d \in [1, p]$ and $\sigma^2_{max}(\bm{W})$ is the largest singular value of $\bm{W}$.
\label{theo3}
\end{theorem}

\begin{proof} 
The proof is shown in Appendix A\ref{A}. 
\end{proof}

The theorem stated above indicates that the proposed method can map nodes with the same label to an area centered around the expectation in the embedding space. This holds true for any graph in which each node's feature and neighborhood pattern are sampled from distributions that depend on the node label.

\begin{table*}[t]
  \centering
  \caption{Accuracy (\%) on the eight datasets for the node classification. The best result is bold, and the second best is underlined.}
  \label{tab1}
  \begin{adjustbox}{max width=\textwidth}
  \begin{tabular}{cccccccccc}
    \toprule
    \multirow{2}{*}{\textbf{Method}} & \multicolumn{9}{c}{\textbf{Datasets}}
    \\
    \cline{2-10}
    &Cora&CiteSeer&PubMed&Computers&Photo&CS&Physics&WikiCS&Avg.\\
    \hline
   {GCN} &81.6{$\pm$0.2} &70.3{$\pm$0.4} &79.3{$\pm$0.2} &84.5{$\pm$0.3} &91.6{$\pm$0.3} &93.1{$\pm$0.3} &93.7{$\pm$0.2} &{73.0{$\pm$0.1}} &83.4\\
   {GAT} &83.1{$\pm$0.3} &72.4{$\pm$0.3} &79.5{$\pm$0.1} &85.8{$\pm$0.1} &91.7{$\pm$0.2} &89.5{$\pm$0.3} &93.5{$\pm$0.3} &{72.6{$\pm$0.3}} &83.5 \\
   {GWNN} &82.8{$\pm$0.3} &71.8{$\pm$0.2} &79.9{$\pm$0.3} &85.6{$\pm$0.2} &92.5{$\pm$0.1} &92.6{$\pm$0.4} &92.7{$\pm$0.2} &72.8{$\pm$0.3} &83.8\\
   {GCNII} &85.5{$\pm$0.4} &73.4{$\pm$0.6} &80.4{$\pm$0.3} &87.6{$\pm$0.4} &92.7{$\pm$0.2} &92.8{$\pm$0.4} &93.5{$\pm$0.2} &74.7{$\pm$0.3} &85.1\\
    \hline
   {GMI} &83.3{$\pm$0.2} &72.6{$\pm$0.2} &79.8{$\pm$0.4} &82.2{$\pm$0.1} &90.7{$\pm$0.2} &92.6{$\pm$0.2} &{94.3{$\pm$0.4}} &74.9{$\pm$0.2} &83.8\\		
   {MVGRL} &83.1{$\pm$0.2} &72.3{$\pm$0.5} &80.3{$\pm$0.5} &87.5{$\pm$0.1} &91.7{$\pm$0.1} &92.1{$\pm$0.3} &{95.1{$\pm$0.2}} &77.5{$\pm$0.1} &84.9\\
   {GCA-SSG} &83.9{$\pm$0.4} &{73.1{$\pm$0.3}} &81.3{$\pm$0.4} &{88.4{$\pm$0.3}} &89.5{$\pm$0.1} &92.4{$\pm$0.1} &93.4{$\pm$0.2} &78.2{$\pm$0.3} &85.0\\
   {GRADE} &{84.0{$\pm$0.3}} &72.4{$\pm$0.4} &{82.7{$\pm$0.3}} &84.7{$\pm$0.1} &{92.6{$\pm$0.1}} &92.7{$\pm$0.4} &93.7{$\pm$0.2} &78.1{$\pm$0.2} &85.1\\
   {AF-GCL} &83.1{$\pm$0.1} &{71.9{$\pm$0.4}} &{79.0{$\pm$0.7}} &\textbf{89.6{$\pm$0.2}} &{92.5{$\pm$0.3}} &{92.0{$\pm$0.1}} &\underline{95.2{$\pm$0.2}} &79.0{$\pm$0.5} &{85.3}\\
   {MA-GCL} &83.3{$\pm$0.4} &73.6{$\pm$0.4} &{83.5{$\pm$0.7}} &{88.8{$\pm$0.3}} &\underline{93.8{$\pm$0.3}} &{92.5{$\pm$0.4}} &{94.8{$\pm$0.5}} &{78.7{$\pm$0.5}} &{86.1}\\
   {GraphMAE} &84.2{$\pm$0.4} &73.4{$\pm$0.4} &{81.1{$\pm$0.4}} &{89.5{$\pm$0.1}} &{93.2{$\pm$0.1}} &{92.7{$\pm$0.2}} &{94.3{$\pm$0.4}} &{78.9{$\pm$0.2}} &{85.9}\\
   {MaskGAE} &84.3{$\pm$0.4} &73.8{$\pm$0.8} &{83.6{$\pm$0.5}} &{89.5{$\pm$0.1}} &{93.3{$\pm$0.1}} &{92.7{$\pm$0.5}} &{94.1{$\pm$0.4}} &{78.4{$\pm$0.2}} &{86.2}\\
    \hline
   {ASWT-SGNN(Semi)} &\textbf{88.1{$\pm$0.4}} &\textbf{81.5{$\pm$0.3}} &\textbf{85.2{$\pm$0.6}} &\underline{89.4{$\pm$0.4}} &\textbf{93.8{$\pm$0.2}} &\underline{93.2{$\pm$0.3}} &94.9{$\pm$0.4} &\textbf{79.8{$\pm$0.5}} &\textbf{88.2}\\
   {ASWT-SGNN(CL)} &\underline{86.2{$\pm$0.6}} &\underline{73.9{$\pm$0.1}} &\underline{84.9{$\pm$0.3}} &89.2{$\pm$0.3} &{93.5{$\pm$0.2}}&\textbf{93.5{$\pm$0.3}} &\textbf{95.4{$\pm$0.3}} &\underline{79.5{$\pm$0.3}} &\underline{87.0}\\
    \bottomrule
  \end{tabular}
  \end{adjustbox}
\end{table*}

\section{Experiments}
\subsection{Experimental Setting}
\textbf{Datasets}\quad We evaluate the approach on eight benchmark datasets, which have been widely used in GCL methods. Specifically, citation datasets include Cora, CiteSeer and PubMed \cite{yang2016revisiting}, co-purchase and co-author datasets include Photo, Computers, CS and Physics \cite{suresh2021adversarial}. Wikipedia dataset includes WikiCS \cite{mernyei2020wiki}. 

\noindent \textbf{Baselines}\quad We consider several baseline methods for the node classification task. These include semi-supervised learning methods like GCN \cite{welling2016semi}, GAT \cite{velickovic2018graph} and GCNII \cite{chen2020simple} and wavelet neural network GWNN \cite{xu2019graph}. Furthermore, we evaluate six GCL methods and two graph generation learning methods, which are GMI \cite{peng2020graph}, MVGRL \cite{hassani2020contrastive}, GCA-SSG \cite{zhang2021canonical}, GRADE \cite{wang2022uncovering}, AF-GCL \cite{wang2022augmentation}, MA-GCL \cite{gong2023ma}, GraphMAE \cite{hou2022graphmae}, and MaskGAE \cite{li2023s}. These methods represent state-of-the-art approaches in the field of node classification tasks.

\noindent \textbf{Evaluation protocol.}\quad For the ASWT-SGNN model, node representations are learned unsupervised using a 2-layer model. Following that, a linear classifier is applied as a post-processing step for assessment. The dataset is randomly partitioned, with 20\% of nodes allocated to the training set, another 20\% to the validation set, and the remaining 60\% to the test set. To ensure the robustness of our findings, we conducted the experiments five times for each dataset, each time with different random seeds. The results include both the average accuracy and the corresponding standard deviation. All experiments use PyTorch on a server with four e NVIDIA A40 GPUs (each 48GB memory). ASWT-SGNN utilizes the Adam Optimizer with a learning rate of 0.001. The specific hyperparameters are as follows: the number of sampling points in the spectral domain, $K$, is set to 20, the feature update ratio, $\alpha$, is set to 0.8, and the wavelet terms ratio, $\beta$, is set to 0.4.

\subsection{Wavelet operator Approximation Experiments}
The proposed method of approximating wavelet operators is applied to real-world graph data. We evaluate its performance using two distinct metrics: the similarity between the approximated wavelet filter $g_\theta$ and the precisely computed wavelet filter, and the Mean Absolute Error (MAE) between the approximated wavelet operator $\bm{\Psi}_\theta$ and the actual wavelet operator. Figure 2 presents a specific example showing the accurate and approximate multiscale filters, computed precisely and approximately, demonstrating a significant overlap between them. The MAEs for different scaled training sets are depicted in Figure 3. The experimental results indicate that the proposed wavelet operator approximation method reduces the approximation error in the high-density spectral domain, while avoiding the need for computationally expensive eigen-decomposition.

\begin{figure}[t]
\centering
    \includegraphics[width=0.23\textwidth,height=2.3cm]{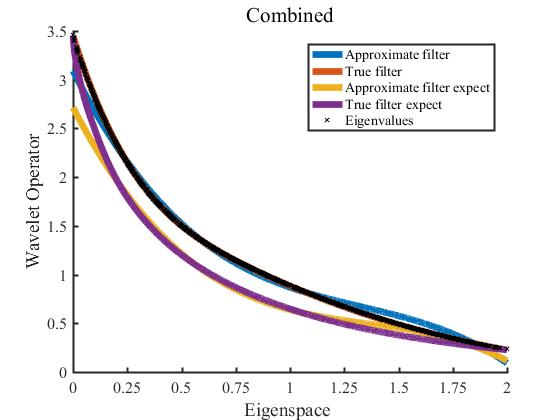}  
    \includegraphics[width=0.23\textwidth,height=2.3cm]{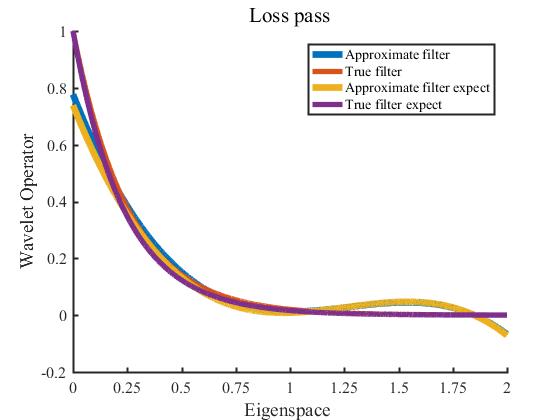}
    \includegraphics[width=0.23\textwidth,height=2.3cm]{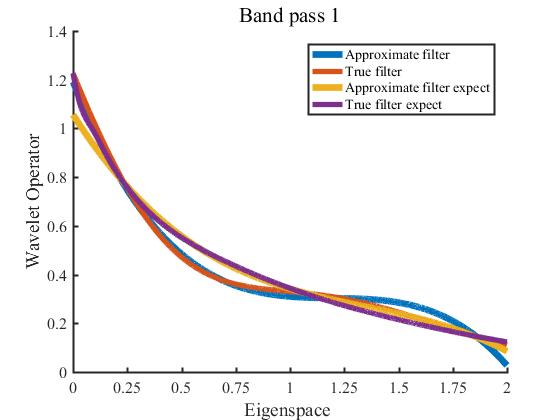}
    \includegraphics[width=0.23\textwidth,height=2.3cm]{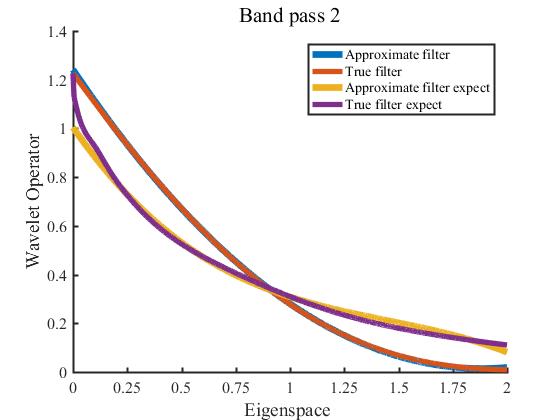}
\caption{Wavelet filters $g_\theta$ for the Cora dataset obtained by actual wavelet and polynomial approximation.}
\label{fig2}
\end{figure}

\begin{figure}[t]
\centering
    \includegraphics[width=0.23\textwidth,height=2.4cm]{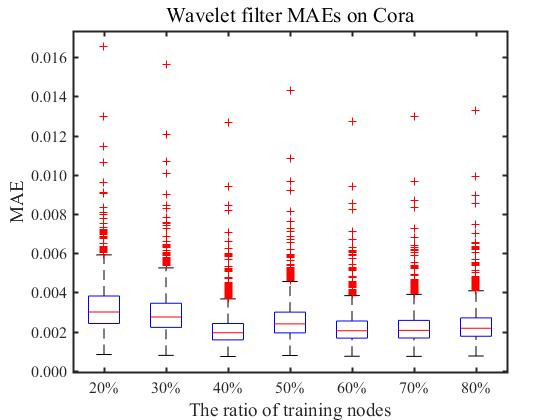}
    \includegraphics[width=0.23\textwidth,height=2.4cm]{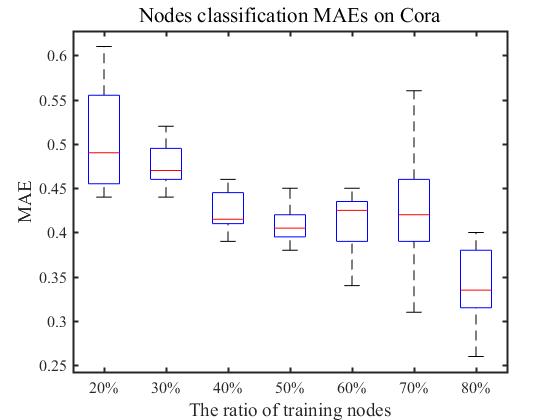}
\caption{MAE between the approximated and the actual wavelet operator at the eigenvalues (left). MAE between the predicted labels and the actual labels of the nodes on the Cora dataset (right).}
\label{fig3}
\end{figure}

\subsection{Node Classification}
Table \ref{tab1} displays the results of node classification accuracy. It is worth noting that our proposed ASWT-SGNN achieves state-of-the-art (SOTA) performance on six out of the eight graphical benchmarks. Specifically, compared to other self-supervised methods across the eight datasets, ASWT-SGNN outperforms the best self-supervised method, MaskGAE, by an average of 0.8\%, and it outperforms the worst self-supervised method, GMI, by 3.2\%. Furthermore, experiments are carried out within a semi-supervised setting, revealing that the proposed method consistently outperforms the best semi-supervised benchmark, GCNII, by an average margin of 3.1\%. Additionally, it surpasses the best self-supervised benchmark, MaskGAE, by an average of 2.0\%. These results further demonstrate the effectiveness of the proposed method ASWT-SGNN in node classification tasks. 

\subsection{Ablation Studies}
\begin{figure}[t]
\centering
    \includegraphics[width=0.23\textwidth,height=2.4cm]{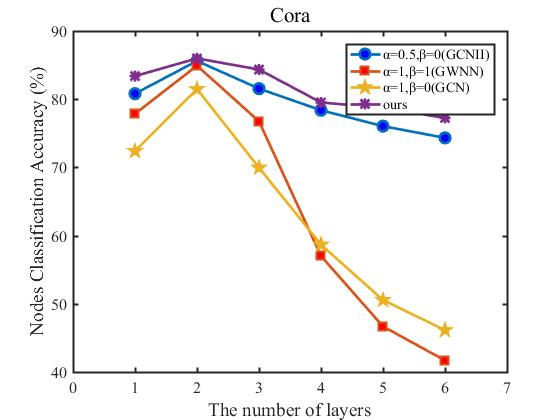} 
    \includegraphics[width=0.23\textwidth,height=2.4cm]{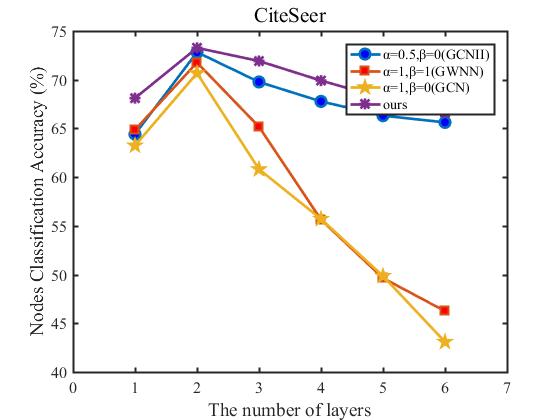}
\caption{Ablation studies. The degree of over-smoothing under various parameter settings.}
\label{fig4}
\end{figure}

The proposed method incorporates two pivotal hyperparameters: $\alpha$ and $\beta$. By adjusting these parameters, ASWT-SGNN can be simplified to its core forms: when $\alpha$ is set to 1 and $\beta$ is set to 0, ASWT-SGNN aligns with GCN; setting both $\alpha$ and $\beta$ to 1 makes ASWT-SGNN behave similarly to GWNN \cite{xu2019graph}; and when $\alpha \neq 1$ and $\beta$ is set to 0, ASWT-SGNN exhibits similarities to GCNII \cite{chen2020simple}. We comprehensively compare ASWT-SGNN and transformation models, including GCN, GWNN, and GCNII. Clear observations can be made from Figure \ref{fig4}: as the number of layers increases, the performance of GCN and GWNN significantly declines. In contrast, the performance of GCNII and ASWT-SGNN remains relatively stable even with more layers stacked. Notably, due to the incorporation of graph wavelet bases, ASWT-SGNN model outperforms GCNII in semi-supervised node classification tasks.

\begin{figure}[t]
\centering
    \includegraphics[width=0.23\textwidth,height=2.5cm]{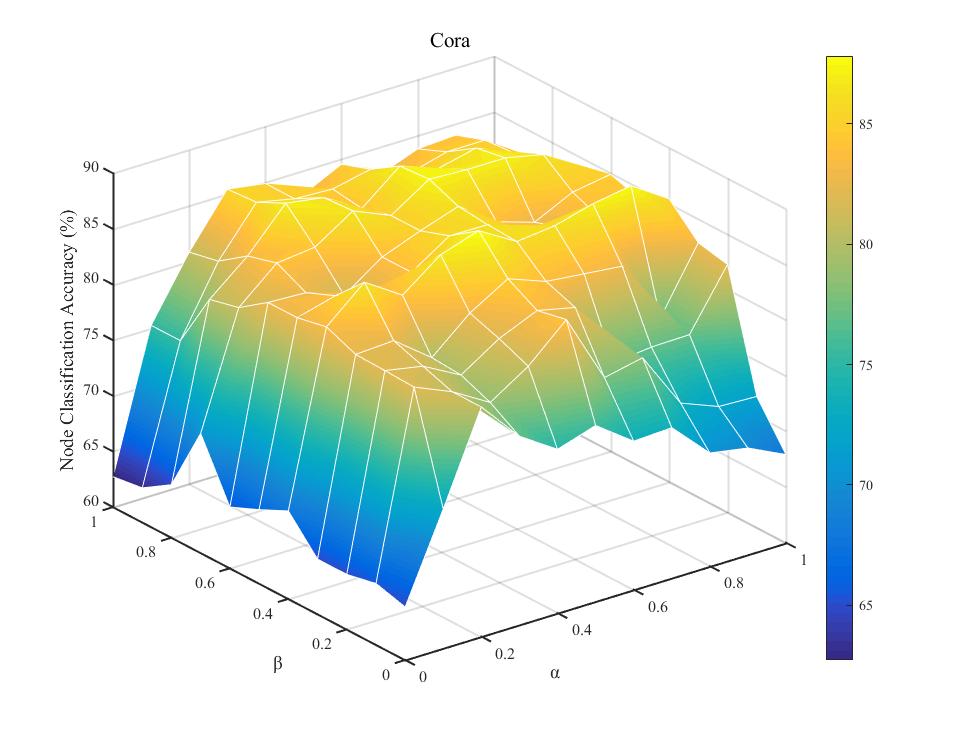}  
    \includegraphics[width=0.23\textwidth,height=2.5cm]{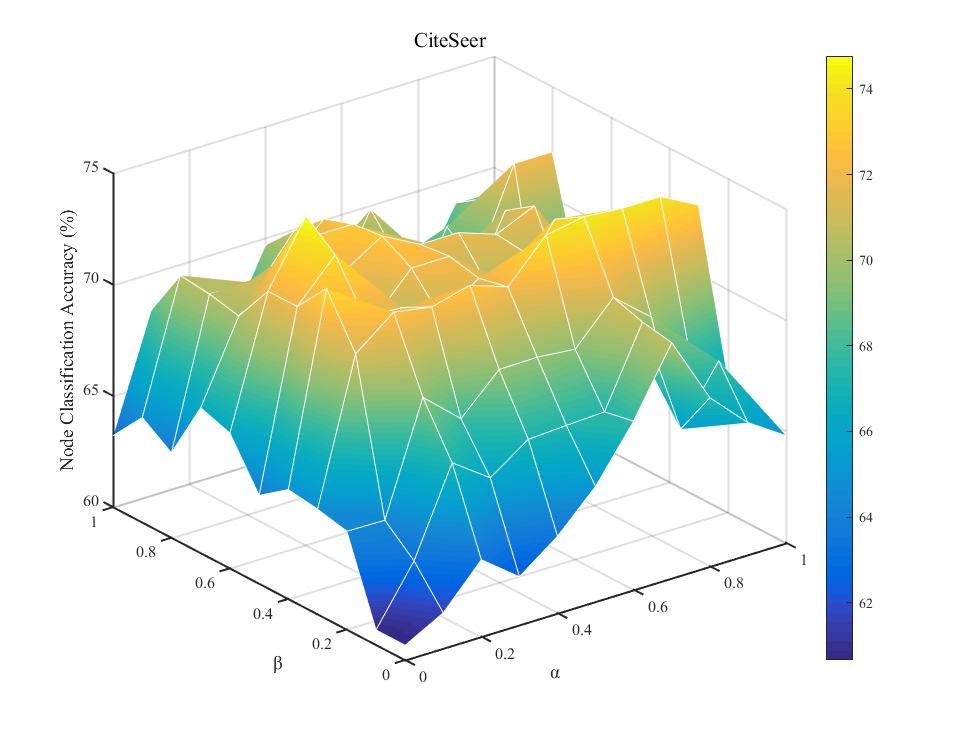}
\caption{Classification accuracy of the proposed method at different parameter settings ($\alpha$ and $\beta$).}
\label{fig5}
\end{figure}

\begin{figure}[t]
\centering
\begin{subfigure}{0.23\textwidth}
    \centering
\includegraphics[width=0.9\textwidth,height=2.3cm]{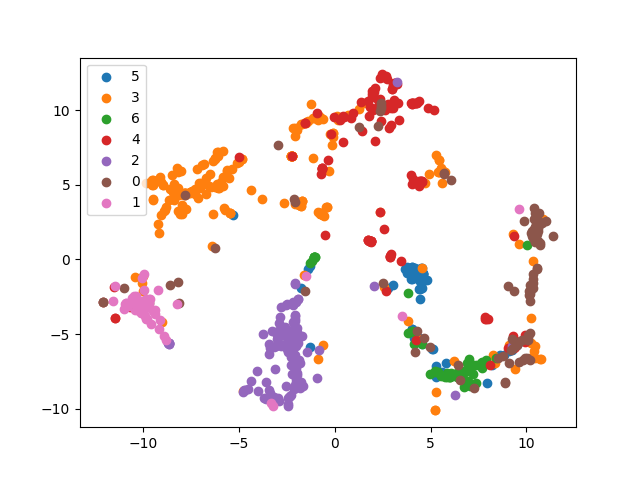}  
    \caption{ASWT-SGNN}
    \label{fig6:subfig3}
\end{subfigure} 
\begin{subfigure}{0.23\textwidth}
    \centering
    \includegraphics[width=0.9\textwidth,height=2.3cm]{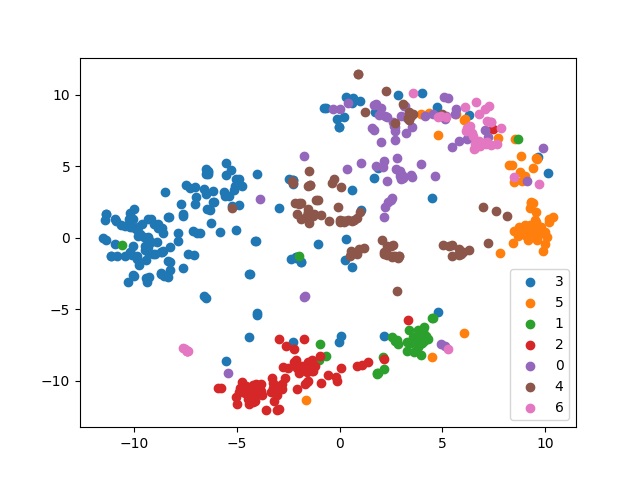}
    \caption{MA-GCL}
    \label{fig6:subfig4}
\end{subfigure}
\caption{Intra-class distance heatmap and node embedding visualization on Cora dataset.}
\label{fig6}
\end{figure}

\subsection{Other Experiments}
To comprehensively investigate the impact of parameters $\alpha$ and $\beta$ on model performance, we systematically vary their values from 0 to 1. The results are presented in Figure \ref{fig5}, illustrating several noticeable trends. When $\alpha$ is set to 0, indicating the exclusion of residual connections, accuracy is significantly decreased. On the other hand, larger values of $\alpha$ result in fixed node representations, leading to lower performance. $\beta$ represents the proportion of the graph wavelet base; if it is tiny, it may not effectively extract local information. Conversely, excessively large values of $\beta$ could result in the neglect of global information, ultimately reducing performance. These findings highlight the complex relationship between parameters $\alpha$ and $\beta$ in shaping the model's performance. This delicate balance allows our model to synthesize local and global information, achieving optimal performance effectively.

Furthermore, we use t-SNE \cite{van2008visualizing} to visualize the node embeddings. Figure \ref{fig6} illustrates that compared to MA-GCL, ASWT-SGNN exhibits more pronounced gaps between different classes. This suggests that ASWL-SGNN captures more detailed class information and clarifies the boundaries between samples from different classes. Further extensive experimental analyses, including sparse, robustness, and uncertainty analyses, are exhaustively presented in Appendix B\ref{B}.

\section{Conclusion}
This paper introduces an adaptive graph wavelet self-supervised neural network called ASWT-SGNN. By utilizing multiple wavelet scales, the model integrates different levels of localization on the graph, enabling the capture of elements beyond the low-frequency ones. To avoid the expensive eigen-decomposition in the spectral domain, the model employs a polynomial approximation of the wavelet operator. Comprehensive experimental results demonstrate the competitiveness of the proposed method against state-of-the-art GCL models on real graph datasets. As our framework applies to all message-passing GNNs and polynomial graph filters, we plan to extend its application to more intricate graph neural architectures. Moreover, we will also consider larger datasets with the increase in GPU resources.

\section{Acknowledgments}
This work is supported in part by the National Natural Science Foundation of China (No.62106259, No.62076234), Beijing Outstanding Young Scientist Program (NO.BJJWZYJH012019100020098), and Beijing Natural Science Foundation (No. 4222029).

\bibliography{aaai24}

\onecolumn  
\newpage    
\twocolumn  

\appendix
\setcounter{theorem}{0} 
\setcounter{lemma}{0}
\section{Appendix A: Details of Methodology}
\label{A}
\subsection{Algorithm of AWST-SGNN}
Algorithm \ref{alg:algorithm} illustrates the detailed steps of deploying AWST-SGNN in an instantiation of GCL.
\begin{algorithm}[h]
\caption{Implementation of AWST-SGNN}
\label{alg:algorithm}
\textbf{Input}: Feature matrix $\bm{X}$; Adjacency matrix $\bm{A}$; Laplacian matrix $\bm{L}$; Encoder $f_\theta$.\\
\textbf{Parameter}: Band-pass filters scales interval $[low_b,high_b]$; Low-pass filters scales interval $[low_l,high_l]$; Number of band-pass filters $L$; Number of sampling points in the spectral domain $K$; Number of Rademacher vectors $n_r$; Feature update ratio $\alpha$; Wavelet terms ratio $\beta$; Learning rate $\eta$.\\
\textbf{Output}: Trained encoder $f_{\theta^*}$ \\
\begin{algorithmic}[1]
\STATE Draw low-pass filter scale $s_0$ from Uniform($low_l$, $high_l$); \\
\STATE Draw band-pass filter scales $\{s_i\}_{i=1}^L$ from Uniform($low_b$, $high_b$); \\
\STATE Draw spectral domain sampling points $\{\xi_i\}_{i=1}^K$ from the linear interval $[0,2]$; \\
\STATE Calculate the spectral density $\bm{\omega}$ according to Eq. (\ref{eq13}); \\
\STATE Initialize encoder $f_\theta$;\\
\STATE \textbf{for} every epoch \textbf{do} \\
\STATE\hspace{\algorithmicindent} Draw two graph augmentations $\bm{o}$ and $\bm{z}$; \\
\STATE\hspace{\algorithmicindent} Calculate the wavelet filters $g_\theta$ according to Eq. (\ref{eq7}); \\
\STATE\hspace{\algorithmicindent} Calculate the approximate wavelet basis $\Psi_\theta$ according to Eq. (\ref{eq14}); \\
\STATE\hspace{\algorithmicindent} View encoding: $\bm{H}_1 = f_\theta(\bm{o}, \Psi_\theta, \bm{A})$, $\bm{H}_2 = f_\theta(\bm{z}, \Psi_\theta, \bm{A})$;\\
\STATE\hspace{\algorithmicindent} Update parameters $\theta$, $\bm{G}$ and $\bm{W}$ by contrastive loss $\mathcal{L}(\bm{H}_1,\bm{H}_2)$.\\
\STATE \textbf{end for} \\
\STATE \textbf{return} Encoder $f_{\theta^*}$

\end{algorithmic}
\end{algorithm}

\subsection{Spectral Density Estimation}
During wavelet operator polynomial approximation, a crucial step involves computing the polynomial coefficients through a weighted least squares approximation. The selection of these weights hinges on their proportionality to the spectral density of the graph, defined as follows:
\begin{equation}
{\omega}(z) = \frac{1}{N}\sum_{j=1}^{N}\{\mathbb{I}(\lambda_{j}= z)\}.
\label{a.eq1}
\end{equation}

Spectral density estimation aims to approximate this function without performing the expensive eigen-decomposition of the graph Laplacian. We opt for the kernel polynomial method to estimate the spectral density function. It involves initially determining an estimate for the cumulative spectral density function $\Omega(z)$ using the formula:
\begin{equation}
\Omega(z)= \frac{1}{N} \sum_{j=1}^{N}\{\mathbb{I}(\lambda_{j} \leq z)\}.
\label{a.eq2}
\end{equation}

For each $\xi$ within the set of $\{\xi_i\}_{i=1}^K$ linearly spaced points across the spectral domain, we aim to ascertain the count of eigenvalues that are less than or equal to $\xi$. Achieving this goal entails approximating the trace of eigen-projection $\bm{P}$ \cite{di2016efficient}, thereby facilitating the computation of an estimate for the eigenvalue counts within the interval $[0, \xi_i]$.
\begin{equation}
\bm{P}=\sum_{\lambda_{j}\in[0,\xi_i]}\bm{u}_{j}\bm{u}_{j}^{T}.
\label{a.eq3}
\end{equation}

The eigenvalues of the projector are restricted to being either zero or one. Consequently, the trace of the projector $\bm{P}$ is equivalent to the count of terms in the Eq.  (\ref{a.eq2}), which corresponds to the number of eigenvalues within the range $[0, \xi_i]$. As a result, the computation of the count of eigenvalues situated within the interval $[0, \xi_i]$ is facilitated by determining the trace of the corresponding projector:
\begin{equation}
tr(\bm{P})=\sum_{j=1}^{N}\{\mathbb{I}(\lambda_{j} \leq \xi_i)\}.
\label{a.eq4}
\end{equation}

If the matrix $\bm{P}$ is explicitly provided, we would be capable of directly calculating its trace and obtaining an exact value for $tr(\bm{P})$. However, the projector $\bm{P}$ is often unavailable in practice. Nonetheless, it is feasible to approximate $\bm{P}$ inexpensively by representing it as either a polynomial or a rational function of the Laplacian matrix $\bm{L}$. In order to achieve this, we can interpret $\bm{P}$ as a step function of $\bm{L}$, specifically:
\begin{equation}
\bm{P} = h(\bm{L}),~\text{where}~h(\lambda)=\left\{\begin{array}{ll}1~&\text{if}~\lambda\le \xi_i\\0~&\text{otherwise}\end{array}\right.
\label{a.eq5}
\end{equation}

The approximation of $h(\lambda)$ is now attainable using a finite sum $\phi(\lambda)$ comprised of Chebyshev polynomials. In this representation, it becomes feasible to estimate the trace of $\bm{P}$ using a stochastic estimator initially introduced by Hutchinson \cite{hutchinson1989stochastic} and subsequently refined by more recent contributions \cite{tang2012probing}. Hutchinson's unbiased estimator employs matrix-vector products exclusively to approximate the trace of a matrix. The foundational principle hinges on applying identically independently distributed (i.i.d.) Rademacher random variables. In this approach, each entry of the randomly generated vectors  $\bm{R}$ assumes either $-1$ or $1$ with equal probability $1/2$. Hutchinson established a lemma demonstrating that $\mathbb{E}(\bm{R}^{\top}\bm{P}\bm{R}) = tr(\bm{P})$. Consequently, an estimation of the trace (\text{tr}(A)) is derived by generating $n_r$ samples of random vectors $\bm{R}_k$, where $k = 1,\cdots,n_r$, and subsequently computing the average of $\bm{R}^{\top}\bm{P}\bm{R}$ across these samples.
\begin{equation}
{tr}(\bm{P})\approx \frac{1}{n_{r}}\sum_{k=1}^{n_{r}}\bm{R}_{k}^{\top}\bm{P}\bm{R}_{k}.
\label{a.eq6}
\end{equation}

In the polynomial filtering approach, the step function $h(\lambda)$ is expanded into a degree $m$ Chebyshev polynomial series:
\begin{equation}
h(\lambda)\approx\phi(\lambda)=\sum_{j=0}^m\gamma_jT_j(\lambda),
\label{a.eq7}
\end{equation}
where $T_j(\lambda)$ are the $j$-degree Chebyshev polynomials, and the coefficients $\gamma_j$ are the expansion coefficients of the step function $h$ which are known to be
\begin{equation}
\gamma_j = \left\{
\begin{array}{ll}
\frac{1}{\pi}\left(\arccos(a)-\arccos(b)\right), & \text{if}~ j=0 \\
\frac{2}{\pi}\left(\frac{\sin(j\arccos(a))-\sin(j\arccos(b))}{j}\right), & \text{if}~ j>0.
\end{array}
\right.
\label{a.eq8}
\end{equation}

As a result, we obtain an expansion of $\bm{P}$ into matrices $T_j(\bm{L})$.
\begin{equation}
h(\bm{L})\approx\phi(\bm{L})=\sum_{j=0}^m\gamma_jT_j(\bm{L}).
\label{a.eq9}
\end{equation}

To mitigate the adverse oscillations near the boundaries (known as Gibbs oscillations) that often arise from the expansion of $h(\lambda)$, a common practice is introducing damping multipliers, referred to as Jackson coefficients. This modification results in the following replacement for Eq. (\ref{a.eq9}):
\begin{equation}
\bm{P} = h(\bm{L}) \approx\phi(\bm{L})=\sum_{j=0}^mg_j^m\gamma_jT_j(\bm{L}).
\label{a.eq10}
\end{equation}

It is worth noting that the matrix polynomial for the conventional Chebyshev approach retains the same form as described above, with the Jackson coefficients $g_j^m$ uniformly set to one. As a result, we will employ the same notation to represent both expansions. The original expression of the Jackson coefficients can be derived using the following formula:
\begin{equation}
\begin{split}
g_j^m= &\frac{\left(1-\frac j{m+2}\right)\sin(\alpha_m)\cos(j\alpha_m)}{\sin(\alpha_m)} \\ &+\frac{\frac1{m+2}\cos(\alpha_m)\sin(j\alpha_m)}{\sin(\alpha_m)},
\end{split}
\label{a.eq11}
\end{equation}
where $\alpha_m=\frac\pi{m+2}$. It is noteworthy that these coefficients can also be expressed in a slightly more concise form:
\begin{equation}
g_j^m=\frac{\sin(j+1)\alpha_m}{(m+2)\sin\alpha_m}+\left(1-\frac{j+1}{m+2}\right)\cos(j\alpha_m).
\label{a.eq12}
\end{equation}

Upon directly substituting the expression for $\phi(\bm{L})$ into the stochastic estimator (Eq. (\ref{a.eq6})), we arrive at the ensuing estimate:
\begin{equation}
\sum_{j=1}^{N}\{\mathbb{I}(\lambda_{j} \leq \xi_i)\}=tr(\bm{P})\approx\frac{1}{n_{v}}\sum_{k=1}^{n_{v}}\left[\sum_{j=0}^{m}g_j^m\gamma_{j}\bm{R}_{k}^{T}T_{j}(\bm{L})\bm{R}_{k}\right].
\label{a.eq13}
\end{equation}

An approximation $\Omega$ to the cumulative spectral density function is obtained using monotonic piece-wise cubic interpolation to interpolate between the estimates at points $\xi_i$.
\begin{equation}
\bm{\Omega}=\mathcal{I}\left(\left\{\left(\xi_{i},\frac{1}{N}\left[\frac{1}{n_{r}}\sum_{k=1}^{n_{r}}\bm{R}_{k}^{\top}\phi({\bm{L}_{sym}})\bm{R}_{k}\right]\right)\right\}_{i=1}^{K}\right).
\label{a.eq14}
\end{equation}

Finally, the approximation $\bm{\omega}$ to the spectral density is derived by differentiating $\bm{\Omega}$ for $\bm{\xi}$.

\subsection{Proof of Lemma \ref{lem1}}
\begin{lemma}
Consider nodes $v_a$ and $v_b$ within a graph, characterized by their similarity in features or labels. Consequently, their $k$-hop neighbors exhibit a one-to-one mapping, specifically $\mathcal{N}_k(b) = \pi(\mathcal{N}_k(a))$. In this context, it holds true that $|\bm{\Psi}_{am} - \bm{\Psi}_{b\pi(m)}| \le 2\epsilon$, where $\bm{\Psi}_{am}$ signifies the wavelet coefficient between nodes $v_a$ and $v_m$.
\label{alem1}
\end{lemma}

\begin{proof}
The definition domain of kernel function $g$ used in wavelet transform in the graph is closed interval $[0,\lambda_N]$. According to the Stone-Weierstrass approximation theorem, a polynomial function can approximate continuous functions on closed intervals. 
\begin{equation}
\begin{split}
\forall \epsilon>0, \exists P(\lambda)=\sum_{j=0}^{m} \alpha_{j} \lambda^{j}:|g(\lambda)-P(\lambda)| \leq \epsilon.
\end{split}
\label{a.eq15}
\end{equation}
Let $r(\lambda)=g(\lambda)-P(\lambda)$, we have:
\begin{equation}
\bm{\Psi}=\left(\sum_{j=0}^{m} \alpha_{j} \bm{L}^{j}\right)+\bm{U} r(\bm{\Lambda}) \bm{U}^{T}.
\label{a.eq16}
\end{equation}

The distribution of wavelet coefficients can be expressed as a polynomial function composed of different terms of the Laplace matrix, and using Newton's binomial theorem to expand the $L^j$ term in the above equation, we have the following form:
\begin{equation}
\begin{split}
\bm{L}^{j}=(\bm{D}-\bm{A})^{j}=\sum_{i=0}^{j} C_{j}^{i} \bm{D}^{i} \bm{A}^{j-i},
\end{split}
\label{a.eq17}
\end{equation}
where $C^i_j$ is the binomial coefficient. If the graph's shortest path between two nodes $v_a$ and $v_b$ is more significant than $j$, we have $\bm{L}^j_{ab}$ = 0.

Then, we discuss the effect of the approximation residue $r(\bm{\Lambda})$ , since the eigenvector matrix $\bm{U}$ is a standard orthogonal matrix and, by virtue of $r(\bm{\lambda}) < \epsilon $, according Cauchy-Schwarz inequality, we obtain:
\begin{equation}
\begin{split}
\left|\delta_{a}^{T} \bm{U} r(\bm{\Lambda}) \bm{U}^{T} \delta_{b}\right|^{2} \leq\left(\sum_{\ell=1}^{N}\left|r\left(\lambda_{\ell}\right)\right|^{2} \bm{U}_{a \ell}^{2}\right)\left(\sum_{\ell=1}^{N} \bm{U}_{b \ell}^{2}\right) \leq \epsilon^{2}.
\end{split}
\label{a.eq18}
\end{equation}

Using Eq. (\ref{a.eq16}), we write the difference between each pair of mapped coefficients $\bm{\Psi}_{am}$ and $\bm{\Psi}_{b\pi(m)}$ in terms of the $k$-th order approximation of the graph Laplacian:
\begin{equation}
\begin{split}
\left|\bm{\Psi}_{m a}-\bm{\Psi}_{\pi(m) b}\right|= \mid \delta_{m} \bm{U}(P(\bm{\Lambda})+r(\bm{\Lambda})) \bm{U}^{T} \delta_{a}- \\
\delta_{\pi(m)} \bm{U}(P(\bm{\Lambda})+r(\bm{\Lambda}))  \bm{U}^{T} \delta_{b}|\leq \\ 
|\left(\bm{U} P(\bm{\Lambda}) \bm{U}^{T}\right)_{m a}-\left(\bm{U} P(\bm{\Lambda}) \bm{U}^{T}\right)_{\pi(m) a} \mid \\
+  \left|\left(\bm{U} r(\bm{\Lambda}) \bm{U}^{T}\right)_{m a}\right|+\left|\left(\bm{U} r(\bm{\Lambda}) \bm{U}^{T}\right)_{\pi(m) b}\right| .
\end{split}
\label{a.eq19}
\end{equation}

Since the $k$-hop neighborhoods around $v_a$ and $v_b$ are identical, the following holds:
\begin{equation}
\begin{split}
\forall m \in \mathcal{N}_{K}(a),\left(\bm{U} P(\bm{\Lambda}) \bm{U}^{T}\right)_{m a} &=\left(\bm{U} P(\bm{\Lambda}) \bm{U}^{T}\right)_{\pi(m) b}, \\
\forall m \notin \mathcal{N}_{K}(a),\left(\bm{U} P(\bm{\Lambda}) \bm{U}^{T}\right)_{m a}&=\left(\bm{U} P(\bm{\Lambda}) \bm{U}^{T}\right)_{\pi(m) a}=0.
\end{split}
\label{a.eq20}
\end{equation}

Therefore, we have $|\bm{\Psi}_{am}-\bm{\Psi}_{b\pi(m)}| \leq 2\epsilon$ according to Eq. (\ref{a.eq20}).
\end{proof}

\subsection{Proof of Theorem \ref{theo3}}
\begin{theorem}
Consider the graph following the above graph assumption and Lemma \ref{lem1}, then the expectation of embedding is given by:
\begin{equation}
\begin{split}
 \mathbb{E}\left[\bm{H}_{i}\right]=\bm{W} \mathbb{E}_{y \sim P\left(y_{i}\right), \bm{x} \sim P_{y}(\bm{x})}[\bm{\Gamma}_i\bm{x}],
\end{split}
\label{eq27}
\end{equation}
where $\bm{\Gamma} = \bm{\Psi} \bm{G} \bm{\Psi}^{\top}$. Thus, with probability at least $1-\delta $ over the distribution for the graph, we have: 
\begin{equation}
\begin{split}
 \left\|\bm{H}_{i}-\mathbb{E}\left[\bm{H}_{i}\right]\right\|_{2} \leq \sqrt{\frac{\sigma_{\max }^{2}(\bm{W}) p \log (2 p / \delta)}{2 \bm{N}\|\bm{\Gamma}_i\bm{x}\|_{\psi_{2}}}},
\end{split}
\label{eq28}
\end{equation}
where the sub-gaussian norms $\|\bm{\Gamma}_i \bm{x}\|_{\psi_{2}} \equiv \min\left\|\bm{\Gamma}_i\bm{x}_{i, d}\right\|_{\psi_{2}}$, $d \in [1, p]$ and $\sigma^2_{max}(\bm{W})$ is the largest singular value of $\bm{W}$.
\label{atheo3}
\end{theorem}

\begin{proof}
We consider a Gaussian mixture model for the node features. For the sake of simplicity, we focus on the binary classification problem. Given the (binary) label $y$ and a latent vector $\boldsymbol{\mu} \sim \mathcal{N}\left(\bm{0}, \bm{I}_p / p\right)$, where the identity matrix $\bm{I}_{p} \in \mathbb{R}^{p \times p}$, the features are governed by: $\bm{x}_{i}=y_{i} \boldsymbol{\mu}+\frac{\bm{q}_{i}}{\sqrt{p}}$. Here, the random variable $\bm{q}_i \in \mathbb{R}^p$ has independent standard normal entries, and $y_i \in \{-1, 1\}$ represents latent classes with abuse of notation. The dimension of the features, denoted as $p$, remains constant. Subsequently, the features of nodes with class $y_i$ follow the same distribution, contingent on $y_i$, i.e., $x_i \sim P_{y_i}(x)$.

When the scale is $\theta$, the sum of wavelet coefficients at node $i$ has the following form:
\begin{equation}
\begin{split}
\sum_{j=1}^{N} \bm{\Psi}_{ij}&=\sum_{j=1}^{N} \sum_{\ell=1}^{N} g_\theta\left( \lambda_{\ell}\right) \bm{U}_{\ell}^{T}(i) \bm{U}_{\ell}(j) \\ &=\sum_{\ell=1}^{N} g_\theta \left(\lambda_{\ell}\right) \bm{U}_{\ell}^{T}(i) \sum_{j=1}^{N} \bm{U}_{\ell}(j).
\end{split}
\label{eq29}
\end{equation}

Because the sum of the elements of the eigenvector corresponding to the non-zero eigenvalues of the Laplace matrix is zero, all the items with $\ell > 1$ can be ignored in the Eq. (\ref{eq29}), and only the items with zero eigenvalues can be retained.

The zero eigenvalues represent the direct current (DC) component of the graph, which corresponds to the same value for each element in the eigenvector and is normalized to $ \bm{U}_{1}=[1 / \sqrt{N}, \ldots, 1 / \sqrt{N}]$, which can be substituted into Eq.(\ref{eq29}) to obtain:
\begin{equation}
\begin{split}
\sum_{j=1}^{N} \bm{\Psi}_{i j}=g(0)=1
\end{split}
\label{eq30}
\end{equation}

Next, we calculate the expectation of aggregated embedding:
\begin{equation}
\begin{split}
\mathbb{E}\left[\bm{H}_{i}\right]&=\mathbb{E}\left[\bm{W} \sum_{j=1}^{N} \bm{\Gamma}_{ij}x_{j}\right] 
 \\ &=\bm{W} \mathbb{E}_{y \sim P\left(y_{i}\right), \bm{x} \sim P_{y}(\bm{\bm{\Gamma}_x})}[\bm{\Gamma}_i\bm{x}],
\end{split}
\label{eq31}
\end{equation}
where $\bm{\Gamma} = \bm{\Psi} \bm{G} \bm{\Psi}^{\top}$. This equation is based on the graph data assumption such that $\bm{x}_{j} \sim P_{y_{i}}(\bm{x})$ for every $j$. Now we provide a concentration analysis. Because each feature $x_i$ is a sub-Gaussian variable, then by Hoeffding's inequality, with probability at least $1-\delta^{'}$ for each $d \in [1, p]$, we have,
\begin{equation}
\begin{split}
 \left|\frac{1}{\bm{N}} \sum_{j}\left(\bm{\Gamma}_i \bm{x}_{j, d}-\mathbb{E}\left[\bm{\Gamma}_i \bm{x}_{j, d}\right]\right)\right| \leq \sqrt{\frac{\log \left(2 / \delta^{\prime}\right)}{2 \bm{N} \left\|\bm{\Gamma}_i \bm{x}_{j, d}\right\|_{\psi_{2}}}},
\end{split}
\label{eq32}
\end{equation}
where $\left\|\bm{\Gamma}_i \bm{x}_{j, d}\right\|_{\psi_{2}}$ is sub-Gaussian norm of $\bm{\Gamma}_i \bm{x}_{j, d}$. Furthermore, because each dimension of $\bm{x}_j$ is mutual independence, thus we define $\left\|\bm{\Gamma}_i \bm{x}_{j}\right\|_{\psi_{2}}=\left\|\bm{\Gamma}_i\bm{x}_{j, d}\right\|_{\psi_{2}}$ Then we apply a union bound by setting $\delta^{\prime}=p \delta$ on the feature dimension $k$. Then with probability at least $1-\delta$ we have, 
\begin{equation}
\begin{split}
 &\left\|\frac{1}{\bm{N}} \sum_{j}\left(\bm{\Gamma}_i\bm{x}_{j, d}-\mathbb{E}\left[\bm{\Gamma}_i\bm{x}_{j, d}\right]\right)\right\|_{2} \\\leq
 &\sqrt{p}\left|\frac{1}{\bm{N}} \sum_{j}\left(\bm{\Gamma}_i\bm{x}_{j, d}-\mathbb{E}\left[\bm{\Gamma}_i\bm{x}_{j, d}\right]\right)\right| \\
 \leq &\sqrt{\frac{p \log (2 p / \delta)}{2 \bm{N}\|\bm{\Gamma}_i\bm{x}\|_{\psi_{2}}}}.
\end{split}
\label{eq33}
\end{equation}

Finally, plug the weight matrix into the inequality,
\begin{equation}
\begin{split}
\left\|\bm{F}_i-\mathbb{E}\left[\bm{F}_i\right]\right\| \leq \sigma_{\max }(\bm{W})\left\|\frac{1}{\bm{N}} \sum_{j}\left(\bm{\Gamma}_i \bm{x}_{j, k}-\mathbb{E}\left[\bm{\Gamma}_i\bm{x}_{j, k}\right]\right)\right\|_{2},
\end{split}
\label{eq34}
\end{equation}
where $\sigma_{max}$ is the largest singular value of the weight matrix.
\end{proof}

\begin{table*}
    \centering
    \begin{tabular}{lllllllll}
        \hline
        Datasets  & Cora & CiteSeer &Pubmed &CS &Photo &Computers &Physics &WikiCS\\
        \hline
        Nodes    &2708 &3327 &19717 &18333 &7650 &13752 &34493 &11701 \\
        Edges    &5429 &4732 &44338 &81894 &119081 &245861 &247962 &216123\\
        Features &1433 &3703 &500 &6805 &745 &767 &8415 &300 \\
        Classes  &5 &6 &3 &15 &8 &10 &5 &10 \\
        \hline
        Density of $\bm{U}$ &99.15\% &95.81\% &99.99\% &100\% &99.66\% &99.55\% &99.83\% &99.20\% \\
        Density of $\bm{\Psi}$ &0.18\% &0.11\% &0.03\% &0.05\% &0.42\% &0.27\% &0.72\% &0.32\%\\
        \hline
    \end{tabular}
    \caption{The Statistics of Datasets.}
    \label{table1}
\end{table*}

\section{Appendix B: Additional Experiments}
\label{B}
\subsection{Datasets}
We evaluate our approach on eight widely used datasets in previous GCL methods, including Cora, Citseeer, Pubmed, Amazon-Photo, Amazon-Computers, Coauthor-CS, Coauthor-Physics and WikiCS. The statistics of datasets are summarized in Table \ref{table1}.
\begin{itemize}
\item \textbf{Cora, CiteSeer, PubMed} \cite{yang2016revisiting}: These datasets are widely used for node classification tasks. They consist of citation networks where nodes represent papers, and edges denote citation relationships.
\item \textbf{Amazon-Computers and Amazon-Photo} \cite{suresh2021adversarial}: These datasets are derived from Amazon's co-purchase relationships. Nodes correspond to products, and edges indicate frequent co-purchases. Each dataset involves product categorization with 10 and 8 classes based on the product category. The node features are bag-of-words representations of product reviews.
\item \textbf{Coauthor-CS and Coauthor-Physics} \cite{suresh2021adversarial}: These academic networks are constructed from co-authorship relationships in the Microsoft Academic Graph. Nodes denote authors, and edges signify co-authored relationships. The datasets involve author classification into 15 and 5 classes based on research fields. Node features are represented as bag-of-words from paper keywords.
\item \textbf{WikiCS} \cite{mernyei2020wiki}: This reference network is constructed from Wikipedia references. Nodes represent computer science articles, and edges denote hyperlinks between articles. Articles are labeled with ten subfields, and their features consist of the average Glove embeddings of all words in the article.
\end{itemize}

\begin{figure*}[t]
\centering
\includegraphics[width=0.24\textwidth,height=3cm]{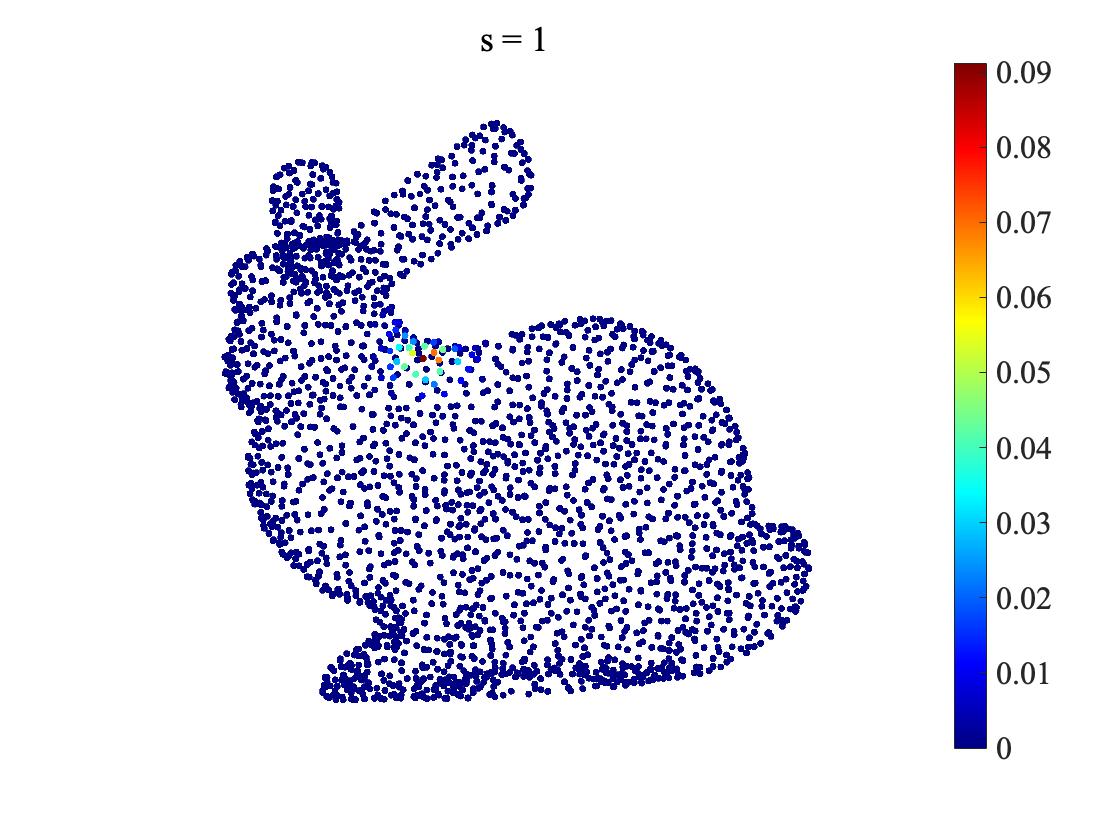}  
\includegraphics[width=0.24\textwidth,height=3cm]{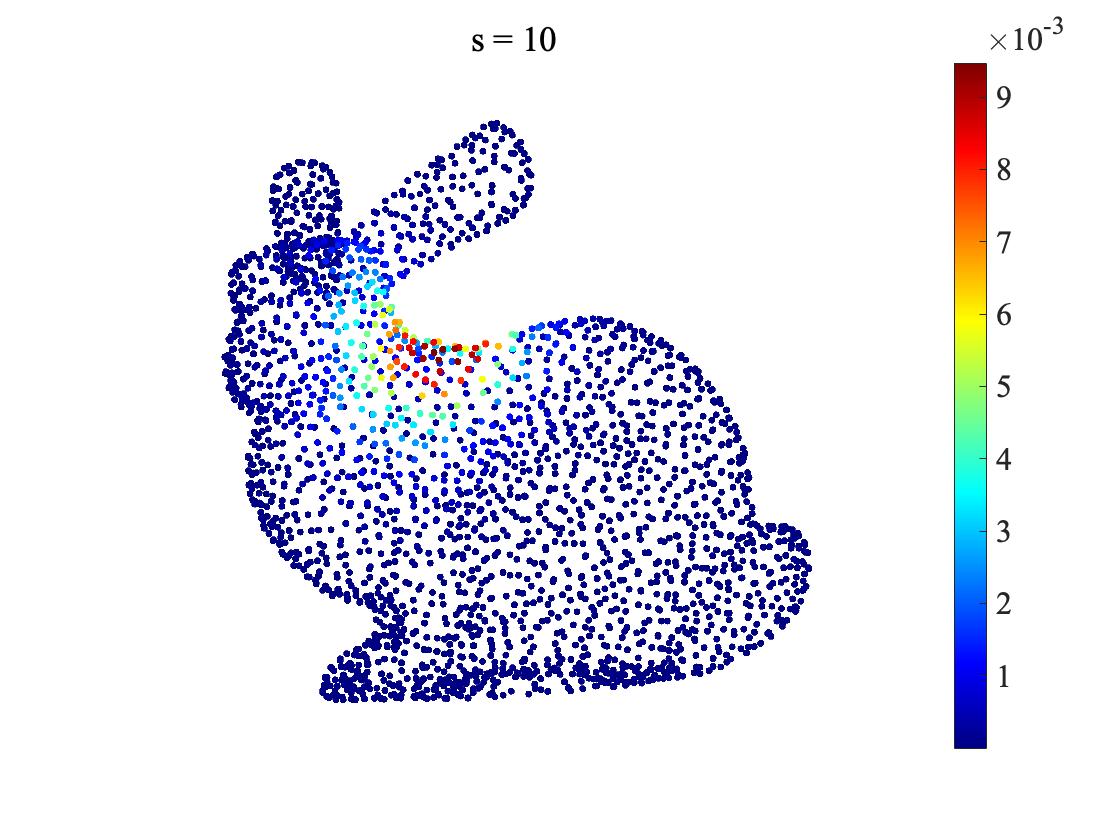}
\includegraphics[width=0.24\textwidth,height=3cm]{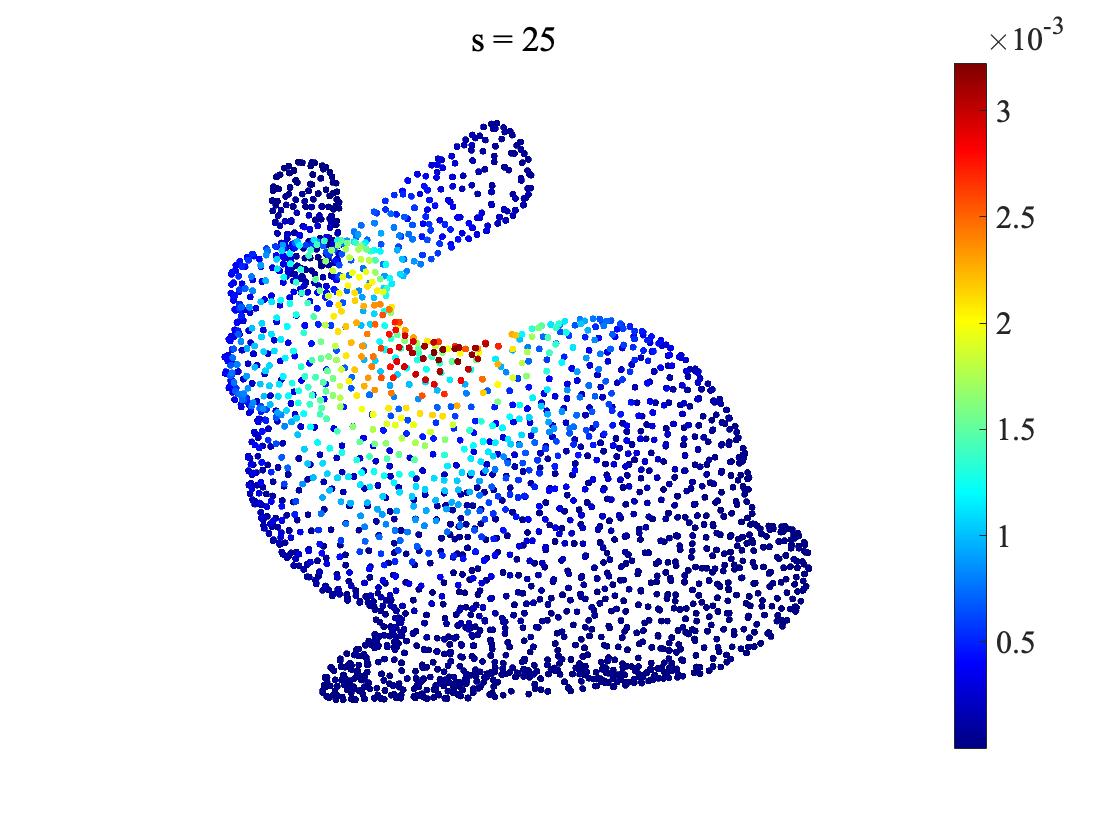}  
\includegraphics[width=0.24\textwidth,height=3cm]{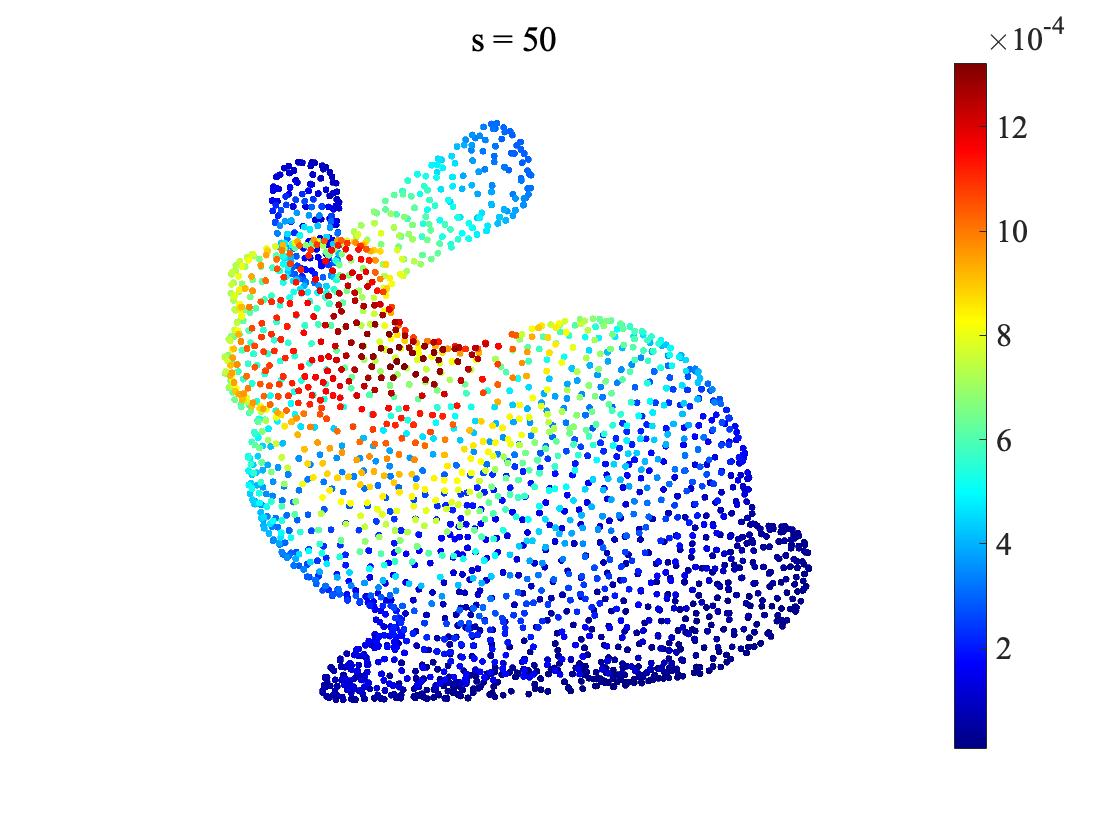}
\caption{The Mexican Hat wavelet transform of a $\delta$ signal on the local node. With different scales, the wavelet can capture different neighborhood information weighted continuously.}
\label{figa1}
\end{figure*}
\begin{figure*}[t]
\centering
\includegraphics[width=0.24\textwidth,height=3cm]{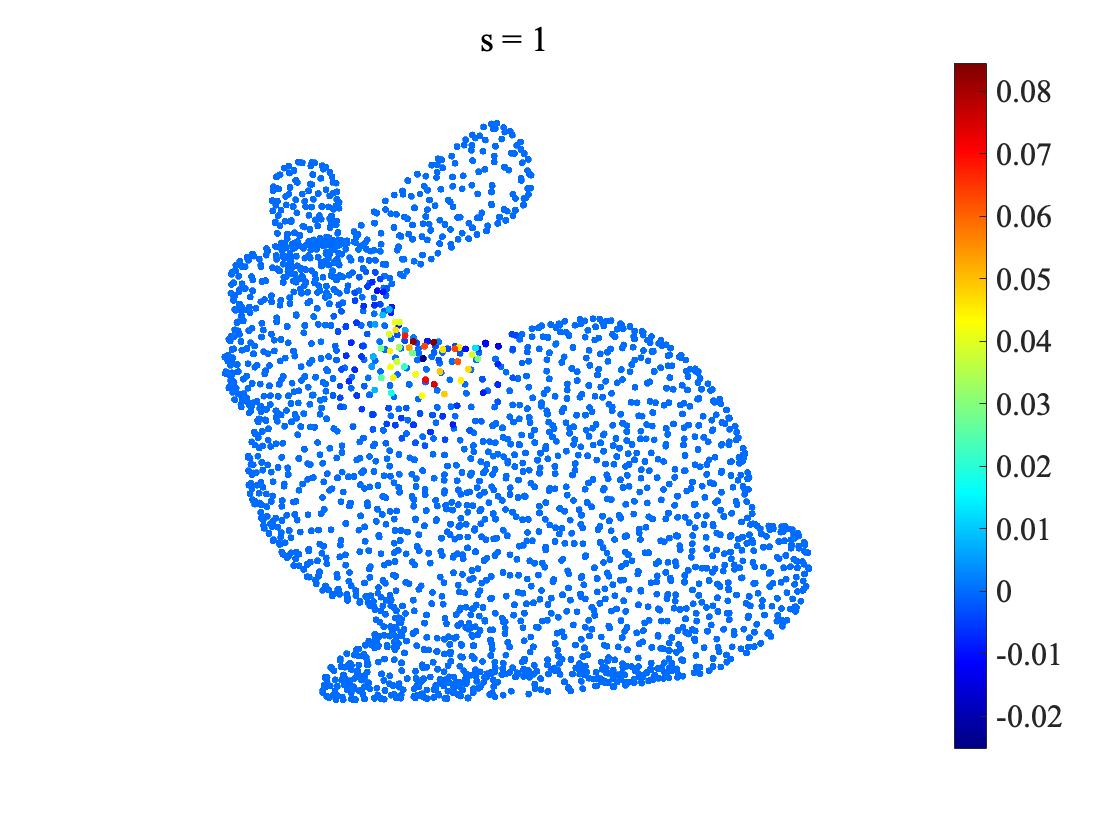}  
\includegraphics[width=0.24\textwidth,height=3cm]{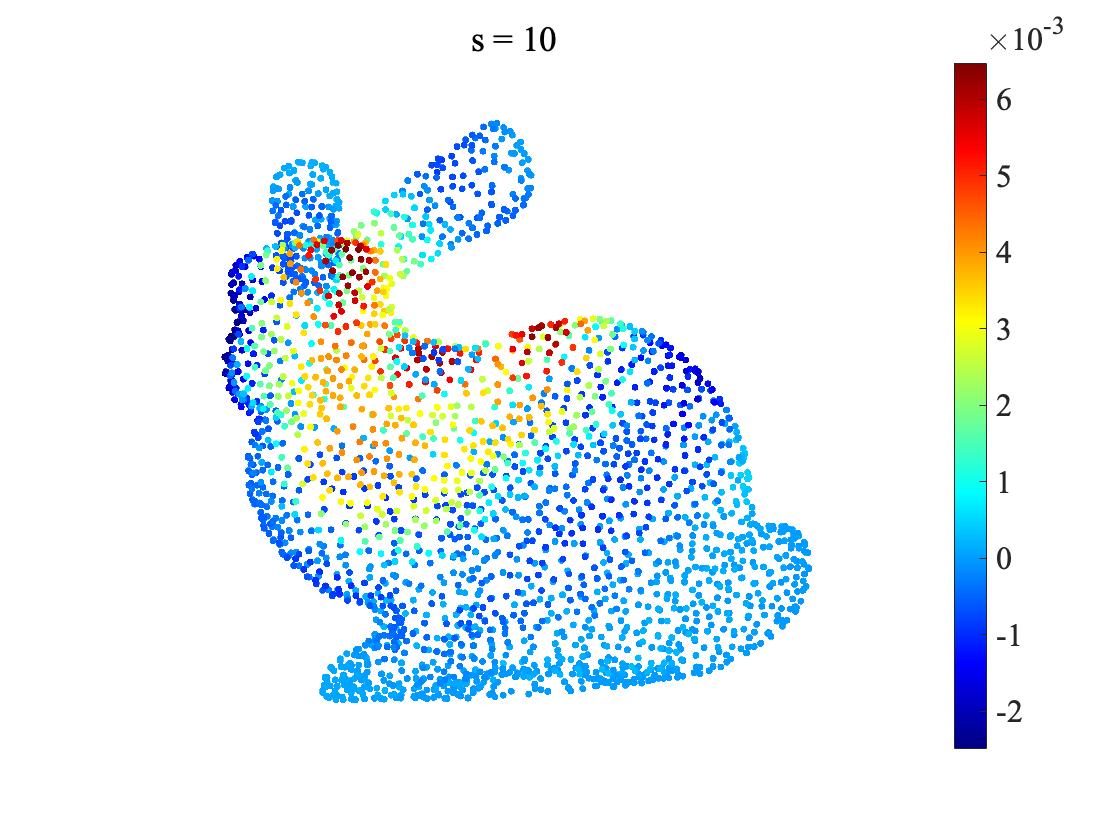}
\includegraphics[width=0.24\textwidth,height=3cm]{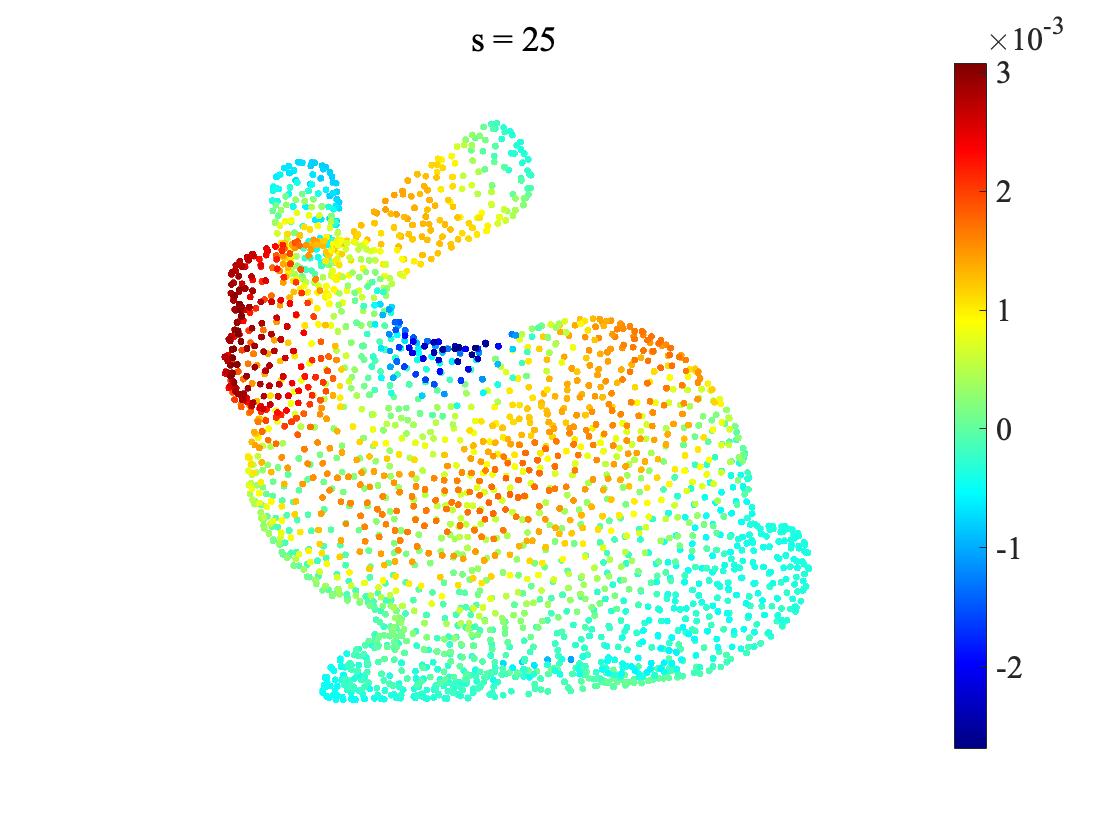}  
\includegraphics[width=0.24\textwidth,height=3cm]{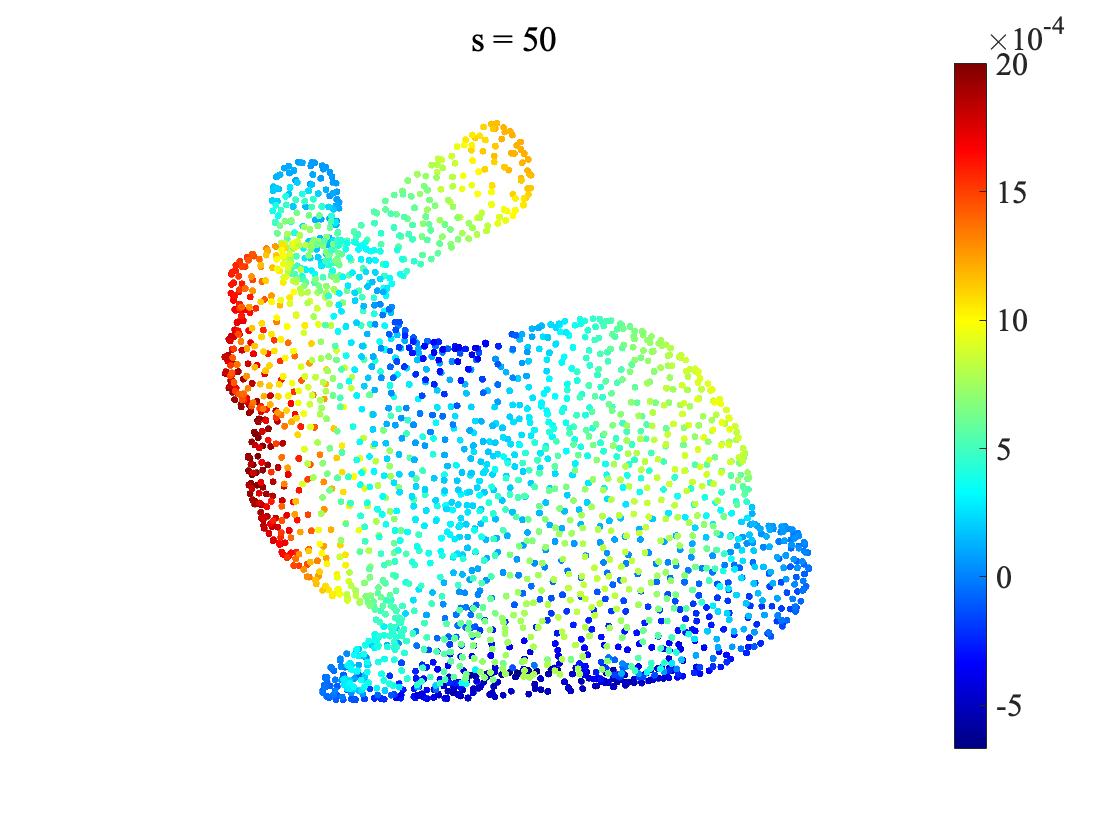}
\caption{The Heat Kernel wavelet transform of a $\delta$ signal on the local node. With different scales, the wavelet can capture different neighborhood information weighted continuously.}
\label{figa2}
\end{figure*}

\begin{figure}[t]
\centering
    \begin{subfigure}{0.23\textwidth}
        \centering
    \includegraphics[width=1\textwidth,height=3cm]{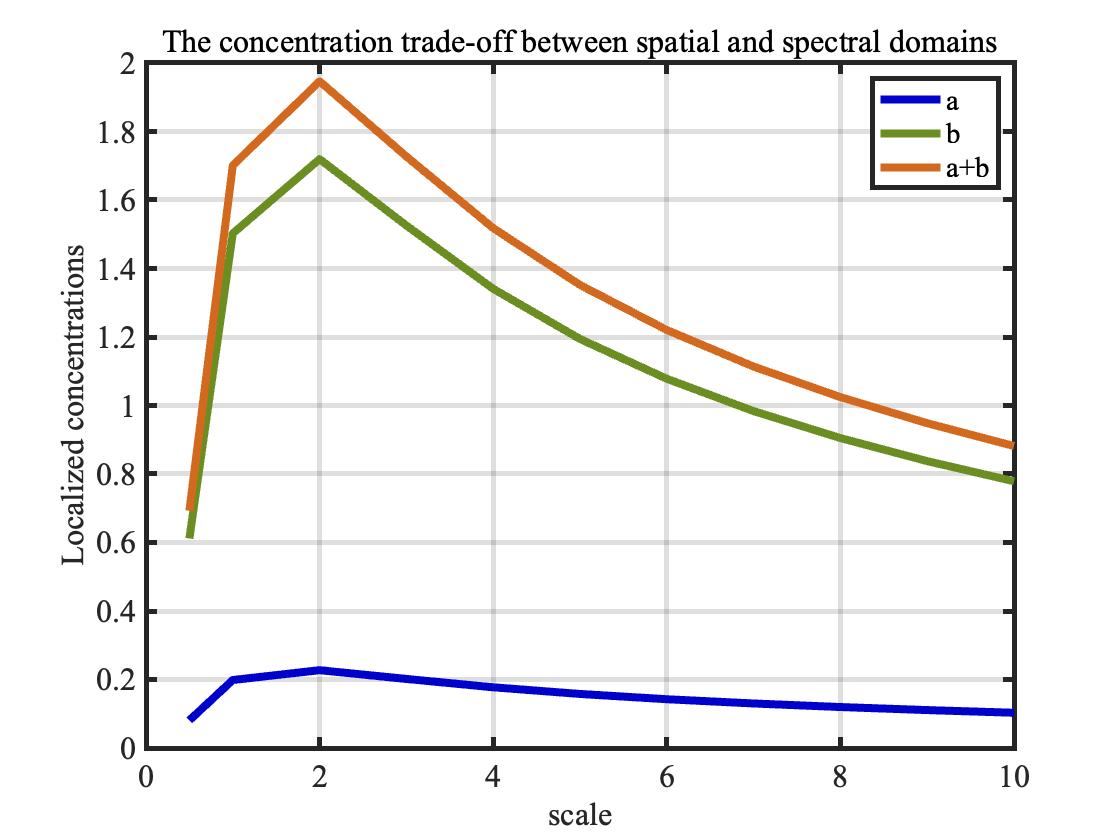}
        \caption{Low-pass filter}
        \label{figa3:subfig1}
    \end{subfigure}
    \begin{subfigure}{0.23\textwidth}
    \centering
    \includegraphics[width=1\textwidth,height=3cm]{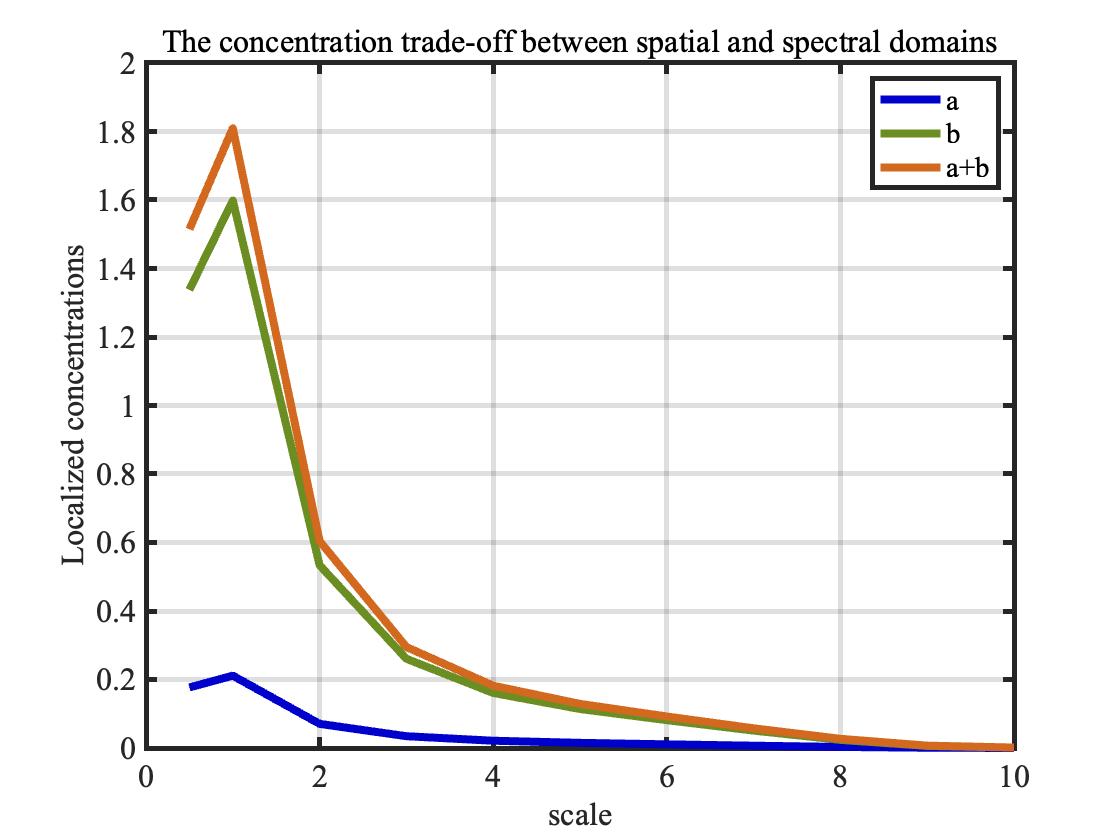}
        \caption{{Band-pass filter}}
        \label{figa3:subfig2}
    \end{subfigure}
    \begin{subfigure}{0.23\textwidth}
        \centering
        \includegraphics[width=1\textwidth,height=3cm]{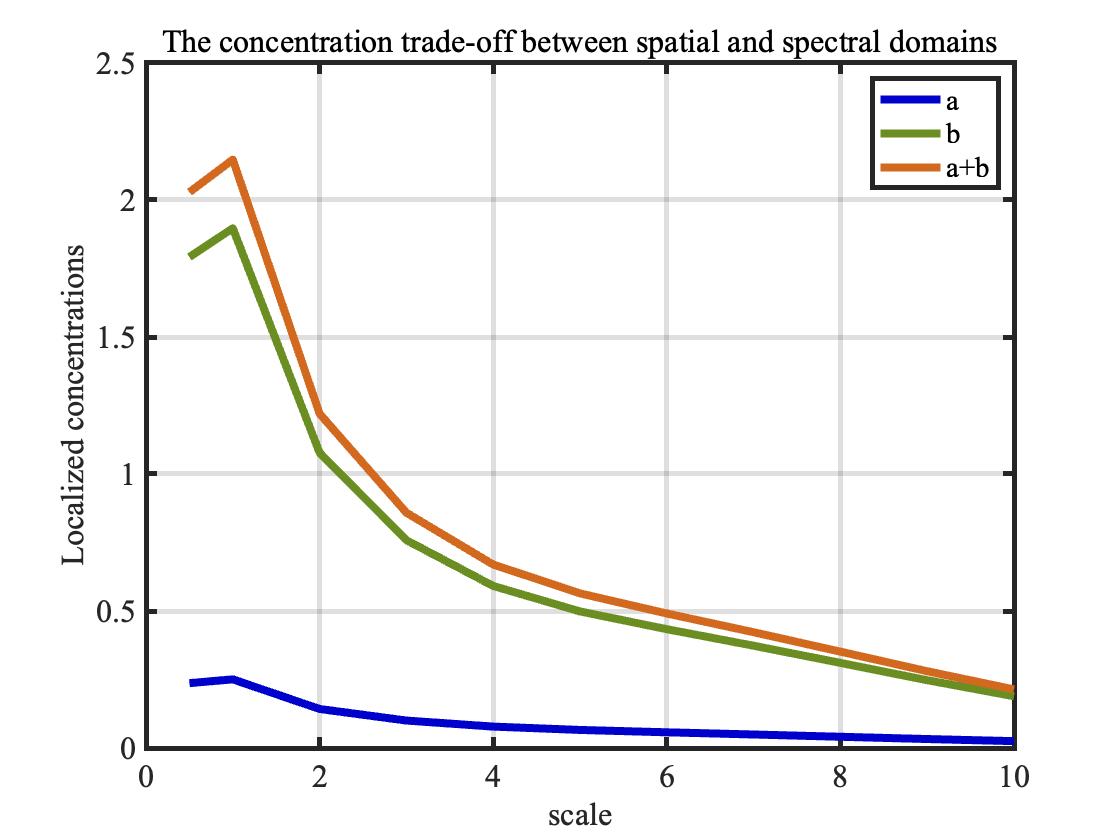}
        \caption{Combined}
        \label{figa3:subfig3}
    \end{subfigure}
\caption{Illustrations of the spatial and spectral domains trade-off in wavelet basis with different scaling parameters $s$.}
\label{figa3}
\end{figure}

\subsection{Baseline}
To ensure consistency, we rely on the code provided by the original authors for all baseline methods, carefully following their recommendations when setting the respective hyperparameters. When specific guidance is lacking, we exercise our judgment to make appropriate adjustments.
\begin{itemize}
    \item GCN: {https://github.com/tkipf/gcn}
    \item GAT: {https://github.com/PetarV-/GAT}
    \item GWNN: {https://github.com/Eilene/GWNN}
    \item GCNII: {https://github.com/chennnM/GCNII}
    \item GMI: {https://github.com/zpeng27/GMI}
    \item MVGRL: {https://github.com/kavehhassani/mvgrl}
    \item GCA-SSG: {https://github.com/hengruizhang98/CCA-SSG}
    \item GRADE: {https://github.com/BUPT-GAMMA/Uncovering-the-Structural-Fairness-in-Graph-Contrastive-Learning}
    \item AF-GCL: {https://github.com/Namkyeong/AFGRL}
    \item MA-GCL: {https://github.com/GXM1141/MA-GCL}
\end{itemize}

\subsection{Parameter Setting}
We furnish a comprehensive account of the hyperparameter settings for all eight datasets. The specifics are outlined below:
\begin{itemize}
    \item Number of band-pass filters: $L$=3
    \item Number of scales: 5
    \item Number of sampling points in the spectral domain: $K$=20
    \item Number of Rademacher vectors: $n_r$=10
    \item The ratio of feature updates: $\alpha$=0.8
    \item The ratio of the graph wavelet terms: $\beta$=0.4
    \item Feature discard ratio: $f_d$=0.2
    \item Order of polynomial approximation: $m$=3
    \item Band-pass filters scales interval: [0,5]
    \item Low-pass filters scales interval: [4,6]
    \item Learning rate: $\eta$=0.001
    \item Hidden size: 256
    \item Projector size: 128
    \item Activation function: ReLU
\end{itemize}
\subsection{Sparsity Analysis}
We computed the sparsity of the Fourier transform's eigenvectors and the wavelet transform's coefficient matrix across the eight datasets. The findings in Table \ref{table1} demonstrate a significant improvement in sparsity for the wavelet transform, with a nearly one hundred-fold increase compared to the Fourier transform. It suggests that the graph wavelet transform exhibits superior computational efficiency over the graph Fourier transform.

\subsection{Uncertainty analysis (Localization analysis)}
To explore the trade-off between spatial and spectral localizations, we can start with an impulse signal $\delta_i$ perfectly localized at the focal node $v_i$. Theoretically, as $s$ approaches 0, the wavelet transform becomes the identity matrix, allowing us to recover the original signal with perfect graph concentration. On the other hand, as $s$ approaches infinity, the graph signal is completely mapped to the lowest frequency, sacrificing graph localization in favor of frequency band concentrations.

For example, let us consider a Stanford Bunny graph (Figure \ref{figa1} and Figure \ref{figa2}). The degree of diffusion for Heat Kernel wavelets and Mexican-Hat wavelets with $s=10$ is much smaller than $s=50$. By choosing $s=10$, we achieve better frequency localization at the expense of graph localization. Using the notation from Eq. (\ref{eq3}) in the body of the paper, we employ a graph limiting operator that covers the direct neighborhood of the focal vertex, denoted as $|\mathcal{S}|=6$, resulting in a concentration in the graph domain.

Figure \ref{figa3} depicts the trade-off between the spatial and spectral domains of different wavelet bases at various scale parameters using the same dataset as shown in Figure \ref{figa1}. Further investigation reveals that a more balanced choice of scale parameter would be in the $[1,3]$ range. Within this range, we can significantly sacrifice a small portion of graph localization to improve frequency localization. In practical scenarios, as demonstrated in the experiment section, the selection of the scale parameter depends on the connectivity of the underlying graph topology and the distribution of training labels in both the spatial and spectral domains. This trade-off between spatial and spectral domains also sheds light on the fundamental limitations of graph signal aggregations. Previous studies have discovered that stacking multiple layers of GCN can lead to over-smoothing problems. From the perspective of the uncertainty principle, deeper networks result in a broader spread of signals in the graph domain, which limits the frequency band's width. This leads to the dominance of low-frequency signals, also known as smooth signals. However, we can precisely control this trade-off with wavelet bases and achieve better results across different datasets.
\begin{equation}
a=\frac{\left \|\bm{B}\bm{U}g_s(\bm{\Lambda})\bm{U}^{\top}\delta_i  \right \| ^2_2}{\left \|\bm{U}g_s(\bm{\Lambda})\bm{U}^{\top}\delta_i \right \| ^2_2}.
\end{equation}

We solve for the maximal $b$ for the given $a$, the maximal possible concentration in the spectral domain becomes \cite{tsitsvero2016signals}:
\begin{equation}
b=a\lambda_{max}+\sqrt{(1-a^2)(1-\lambda_{max}^2)}.
\end{equation}

\begin{figure*}[t]
\centering
    \begin{subfigure}{0.24\textwidth}
        \centering
    \includegraphics[width=1\textwidth,height=4cm]{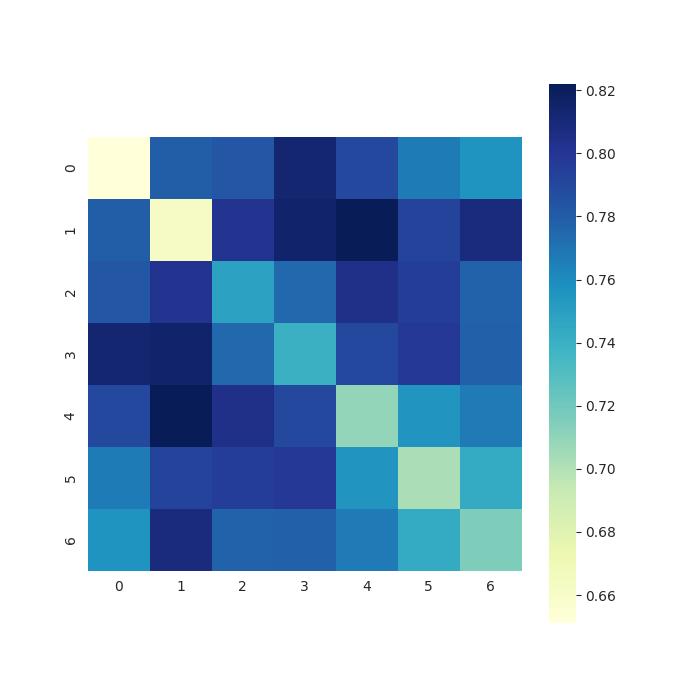}
        \caption{Epoch 1}
        \label{figa5:subfig1}
    \end{subfigure}
    \begin{subfigure}{0.24\textwidth}
    \centering
    \includegraphics[width=1\textwidth,height=4cm]{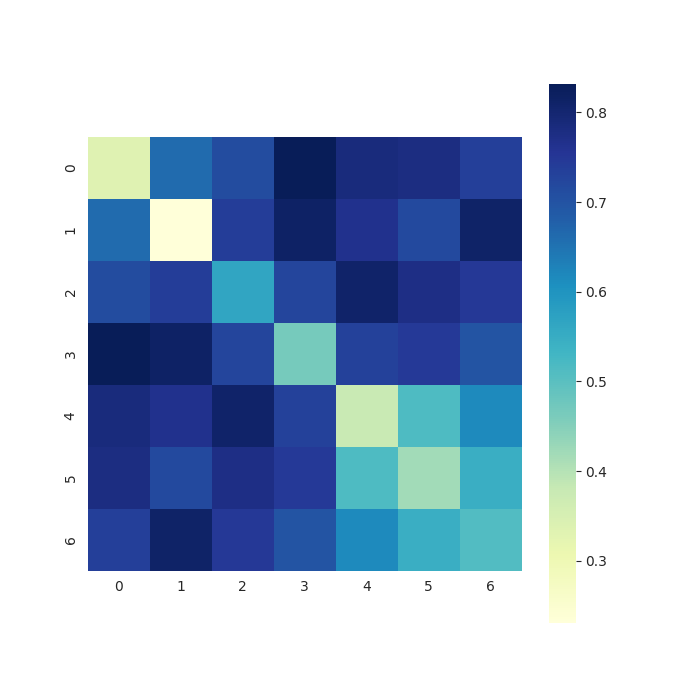}
        \caption{Epoch 100}
        \label{figa5:subfig2}
    \end{subfigure}
    \begin{subfigure}{0.24\textwidth}
        \centering
        \includegraphics[width=1\textwidth,height=4cm]{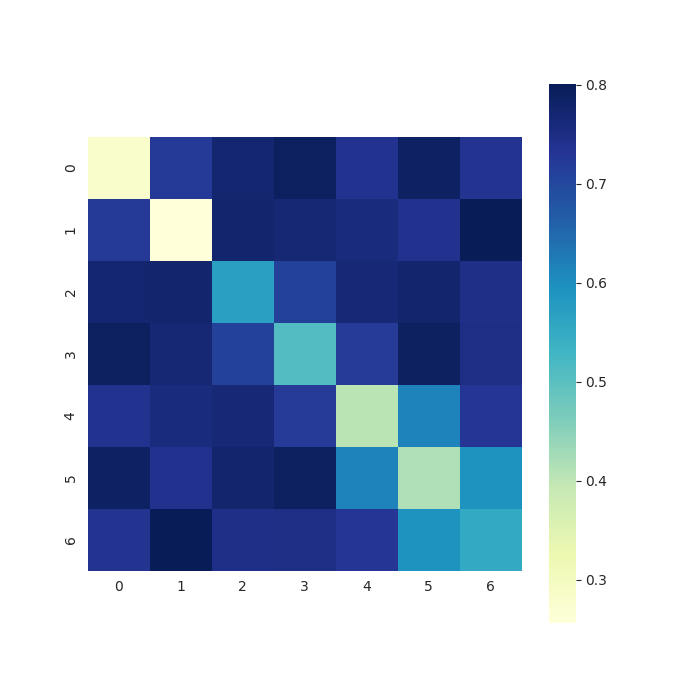}
        \caption{Epoch 300}
        \label{figa5:subfig3}
    \end{subfigure}
    \begin{subfigure}{0.24\textwidth}
        \centering
        \includegraphics[width=1\textwidth,height=4cm]{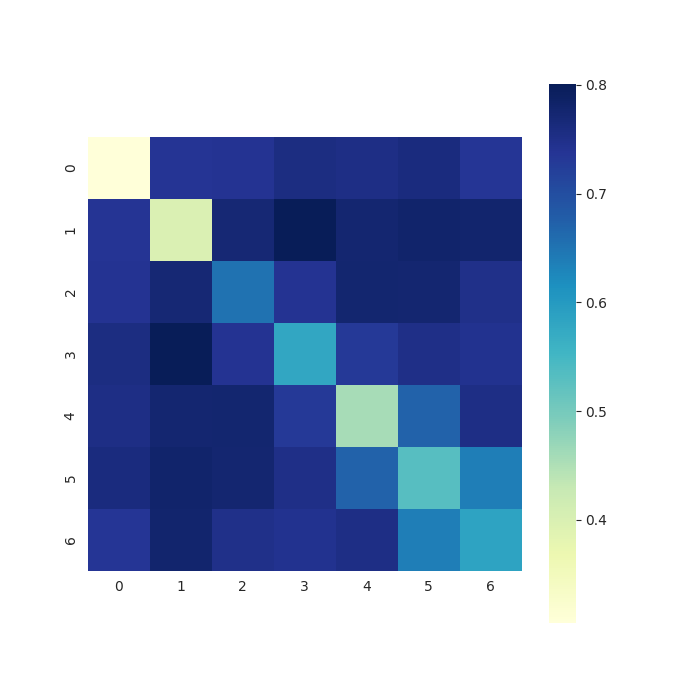}
        \caption{Epoch 500}
        \label{figa5:subfig4}
    \end{subfigure}
    \begin{subfigure}{0.24\textwidth}
        \centering
        \includegraphics[width=1\textwidth,height=4cm]{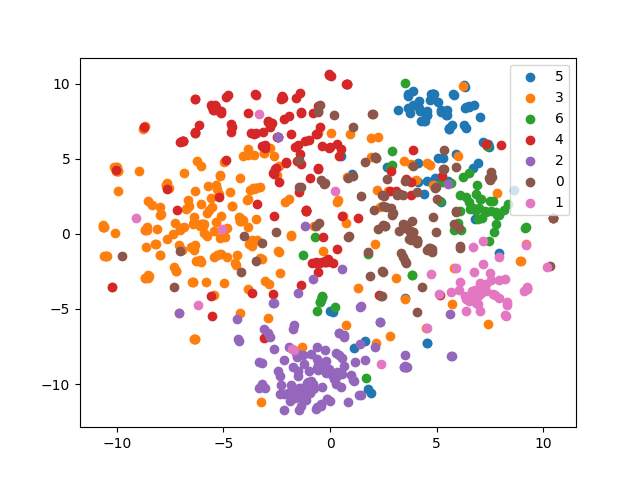}
        \caption{Epoch 1}
        \label{figa5:subfig5}
    \end{subfigure}
    \begin{subfigure}{0.24\textwidth}
        \centering
        \includegraphics[width=1\textwidth,height=4cm]{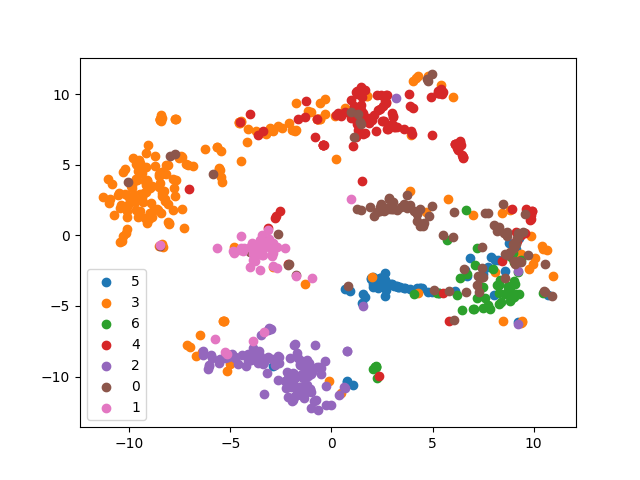}
        \caption{Epoch 100}
        \label{figa5:subfig6}
    \end{subfigure}
    \begin{subfigure}{0.24\textwidth}
        \centering
        \includegraphics[width=1\textwidth,height=4cm]{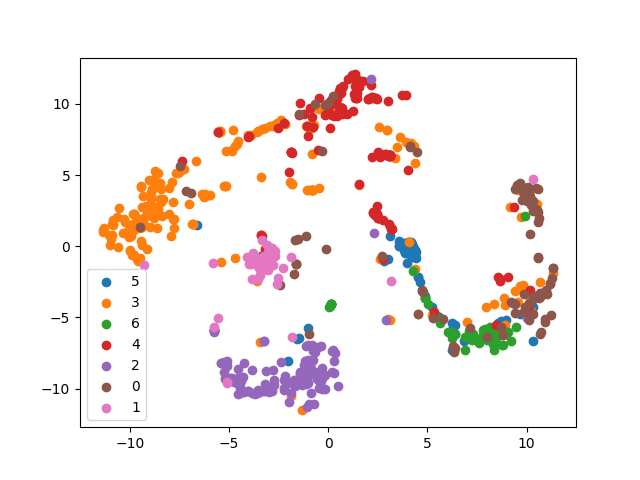}
        \caption{Epoch 300}
        \label{figa5:subfig7}
    \end{subfigure}
    \begin{subfigure}{0.24\textwidth}
        \centering
    \includegraphics[width=1\textwidth,height=4cm]{pic/cora_vis_epoch500.png}
        \caption{Epoch 500}
        \label{figa5:subfig8}
    \end{subfigure}
\label{afig}
\caption{The visualization of learned features of the Cora dataset during training ASWT-SGNN.}
\end{figure*}

\subsection{Graph Classification}
For graph classification, we conduct experiments on three benchmarks: MUTAG, NCI1, and IMDB-BINARY \cite{yanardag2015deep}. Each dataset is a collection of graphs where each graph is associated with a label. 

For the evaluation protocol, after generating graph embeddings with AWST-SGNN’s encoder and readout function, we feed encoded graph-level representations into a downstream classifier to predict the label, and report the mean 10-fold cross-validation accuracy with standard deviation after five runs. 

In addition to the classical graph kernel methods, namely Weisfeiler-Lehman Subtree Kernel (WL) \cite{shervashidze2011weisfeiler} and GNTK \cite{du2019graph}, we compare AWST-SGNN with three supervised methods: GIN \cite{xu2018powerful}, GraphSAGE \cite{hamilton2017inductive} , and DGCNN \cite{wang2019dynamic}, as well as two self-supervised methods: GraphCL \cite{you2020graph} and MVGRL \cite{hassani2020contrastive}. The results are presented in Table \ref{tab3}. It is observed that AWST-SGNN outperforms all the baselines on two datasets, and on the remaining dataset it achieves competitive performance. These results suggest that AWST-SGNN is capable of learning meaningful information and holds promise for graph-level tasks.

\begin{table}[t]
  \centering
  \caption{Accuracy (\%) on the three datasets for the graph classification. The best result is bold, and the second best is underlined.}
  \label{tab3}
  \begin{adjustbox}{max width=0.5\textwidth}
  \begin{tabular}{ccccc}
    \toprule
    \multirow{2}{*}{\textbf{Method}} & \multicolumn{4}{c}{\textbf{Datasets}}
    \\
    \cline{2-5}
    &MUTAG&NCI1&IMDB-BINARY&Avg.\\
    \hline
   {WL} &88.7{$\pm$7.3} &76.7{$\pm$2.1} &72.8{$\pm$3.9} &79.4 \\
   {GNTK} &89.3{$\pm$8.5} &76.8{$\pm$4.9} &\textbf{75.9{$\pm$4.6}} &80.6 \\
   {GIN} &89.4{$\pm$5.6} &\underline{80.0{$\pm$2.4}} &75.1{$\pm$2.8} &\underline{81.5} \\
   {DGCNN} &85.9{$\pm$1.7} &74.1{$\pm$3.1} &69.2{$\pm$3.4} &76.4\\
   {GraphSAGE} &86.6{$\pm$1.5} &74.8{$\pm$2.3} &68.9{$\pm$4.5} &76.7 \\
   
   {GraphCL} &86.8{$\pm$1.3} &77.9{$\pm$2.4} &71.1{$\pm$2.1} &{78.6} \\
   {MVGRL} &\underline{89.4{$\pm$1.9}} &79.1{$\pm$1.7} &74.2{$\pm$2.7} &{80.9} \\
     \hline
   {Ours} &\textbf{89.7{$\pm$5.2}} &\textbf{80.2{$\pm$3.8}} &\underline{75.7{$\pm$3.1}} &\textbf{81.9} \\
    \bottomrule
  \end{tabular}
  \end{adjustbox}
\end{table}

\subsection{Robustness analyses}
\begin{figure}[t]
\centering
    \includegraphics[width=0.23\textwidth,height=3cm]{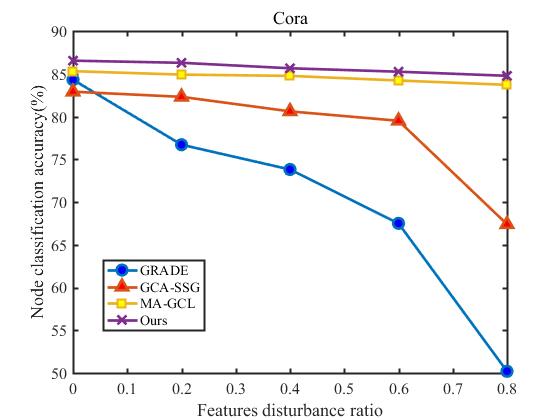}  
    \includegraphics[width=0.23\textwidth,height=3cm]{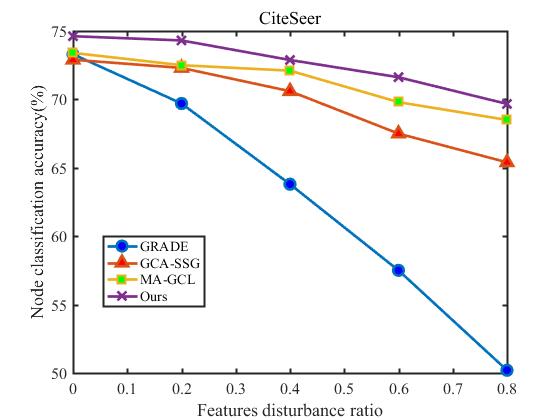}
\caption{Test accuracy when graphs are perturbed by features masking.}
\label{figa4}
\end{figure}

To assess the adaptability of ASWL-SGNN to adversarial graphs, we conducted experiments on the Cora and Citeseer datasets. This evaluation involved introducing random feature masking to the original graph structure, thereby varying the ratio of masked features from 0 to 0.8 to simulate varying degrees of attack intensity. We performed comparisons against GRADE, GCA-SSG, and MA-GCL. The outcomes, illustrated in Figure \ref{figa4}, consistently demonstrate ASWL-SGNN's superior performance over the other methods. Significantly, our method displays more substantial performance enhancements as the feature masking rate increases. This observation suggests that ASWL-SGNN exhibits heightened robustness against severe feature attacks.

\subsection{Visualisation}
For AWST-SGNN, we use heatmaps to show the intra-class distances and inter-class distances during the training period on the Cora dataset, as shown in Figure \ref{figa5:subfig1}-\ref{figa5:subfig4}, where the intra-class distances are shrinking, and the inter-class distances are increasing as the training progresses. Moreover, the t-SNE algorithm \cite{van2008visualizing} is utilized to visualize the learned embeddings during the training process (as shown in Figure \ref{figa5:subfig5}-\ref{figa5:subfig8}). Each distinct color within the visualization corresponds to a different class. As training advances, the disparities between the feature representations of various categories become more discernible.

\end{document}